\documentclass[10pt,twocolumn,letterpaper]{article}

\usepackage{cvpr}              
\definecolor{cvprblue}{rgb}{0.21,0.49,0.74}
\usepackage[pagebackref,breaklinks,colorlinks,allcolors=cvprblue]{hyperref}
\usepackage{wrapfig}
\usepackage{booktabs}
\usepackage{hyperref}
\usepackage[table]{xcolor}
\usepackage{makecell}
\usepackage{url}
\usepackage{graphicx} 
\usepackage{array,multirow}
\usepackage{arydshln}
\usepackage{adjustbox}
\usepackage{subcaption}
\usepackage{caption}
\usepackage{hyperref}
\usepackage{url}
\usepackage[dvipsnames]{xcolor}
\definecolor{mydarkorange}{HTML}{B86046}
\hypersetup{
    colorlinks=true,
    linkcolor=mydarkorange,
    citecolor=codepurple,
    filecolor=mydarkorange,
    urlcolor=mydarkorange
}

\definecolor{codegreen}{rgb}{0,0.6,0}
\definecolor{codegray}{rgb}{0.5,0.5,0.5}
\definecolor{codepurple}{rgb}{0.58,0,0.82}
\definecolor{backcolour}{rgb}{0.95,0.95,0.92}
\definecolor{bg}{rgb}{0.95,0.95,0.95}
\definecolor{mygreen}{rgb}{0,0.6,0}
\def\paperID{*****} 
\def\confName{CVPR}
\def\confYear{2026}

\title{\textit{Saliency-R1}: Enforcing Interpretable and Faithful  Vision-language \\ Reasoning via Saliency-map Alignment Reward}

\author{Shizhan Gong$^1$, Minda Hu$^1$, Qiyuan Zhang$^2$, Chen Ma$^2$, Qi Dou$^{1,}$\thanks{Corresponding author}\\
$^1$The Chinese University of Hong Kong, Hong Kong, China\\
$^2$City University of Hong Kong, Hong Kong, China\\
{\tt\small \{szgong22, mdhu22, qdou\}@cse.cuhk.edu.hk}
{\tt\small \{qyzhang, chenma\}@cityu.edu.hk}
}

\begin{document}
\maketitle
\begin{abstract}
Vision-language models (VLMs) have achieved remarkable success across diverse tasks. However, concerns about their trustworthiness persist, particularly regarding tendencies to lean more on textual cues than visual evidence and the risk of producing ungrounded or fabricated responses. To address these issues, we propose \textbf{Saliency-R1}, a framework for improving the interpretability and faithfulness of VLMs reasoning. Specifically, we introduce a novel saliency map technique that efficiently highlights critical image regions contributing to generated tokens without additional computational overhead. This can further be extended to trace how visual information flows through the reasoning process to the final answers, revealing the alignment between the thinking process and the visual context. We use the overlap between the saliency maps and human-annotated bounding boxes as the reward function, and apply Group Relative Policy Optimization (GRPO) to align the salient parts and critical regions, encouraging models to focus on relevant areas when conduct reasoning. Experiments show Saliency-R1 improves reasoning faithfulness, interpretability, and overall task performance.  The code is available at \url{https://github.com/peterant330/Saliency_R1}.
\end{abstract}

\section{Introduction}
Vision-language models (VLMs) have made impressive strides, demonstrating strong performance on tasks that require joint reasoning over images and text. As a result, VLMs have been widely adopted in various domains, including high-stakes areas such as autonomous driving~\citep{cui2024survey}, finance~\citep{lee2025large}, and medical diagnosis~\citep{gong2025concepts}. Despite their superior performance, the trustworthiness of VLMs has raised concerns. For example,  prior work show that VLMs often over-rely on textual cues, with visual signals playing a comparatively smaller role~\citep{parcalabescu2025do,hao2025can}. Another concern is hallucination~\citep{favero2024multi}, where models produce ungrounded or fabricated content not supported by the image. These issues limit VLMs’ broader deployment in safety-critical settings.

To build trust, VLMs are frequently prompted to generate natural language explanations (NLEs) that describe their reasoning. For instance, Chain-of-Thought (CoT) prompts~\citep{wei2022chain} can be used to elicit step-by-step reasoning before producing a final answer. Recent reasoning-based language models, such as DeepSeek-R1~\citep{guo2025deepseek} and Claude-Sonnet~\citep{anthropic2024claude35}, are explicitly trained to provide detailed reasoning traces. However, these NLEs still have limitations. Multiple studies report inconsistencies between the articulated reasoning process and the final answer~\citep{parcalabescu2024measuring,admoni2025towards,barez2025chain,yu2025explainablechainofthoughtreasoningempirical}, indicating that the provided “thinking” may not faithfully reflect the process that led to the solution. Moreover, the explanations themselves can exhibit problems including misusing visual cues or hallucinating unsupported details.
\begin{figure}[t]
\centering
\includegraphics[width=0.5\textwidth]{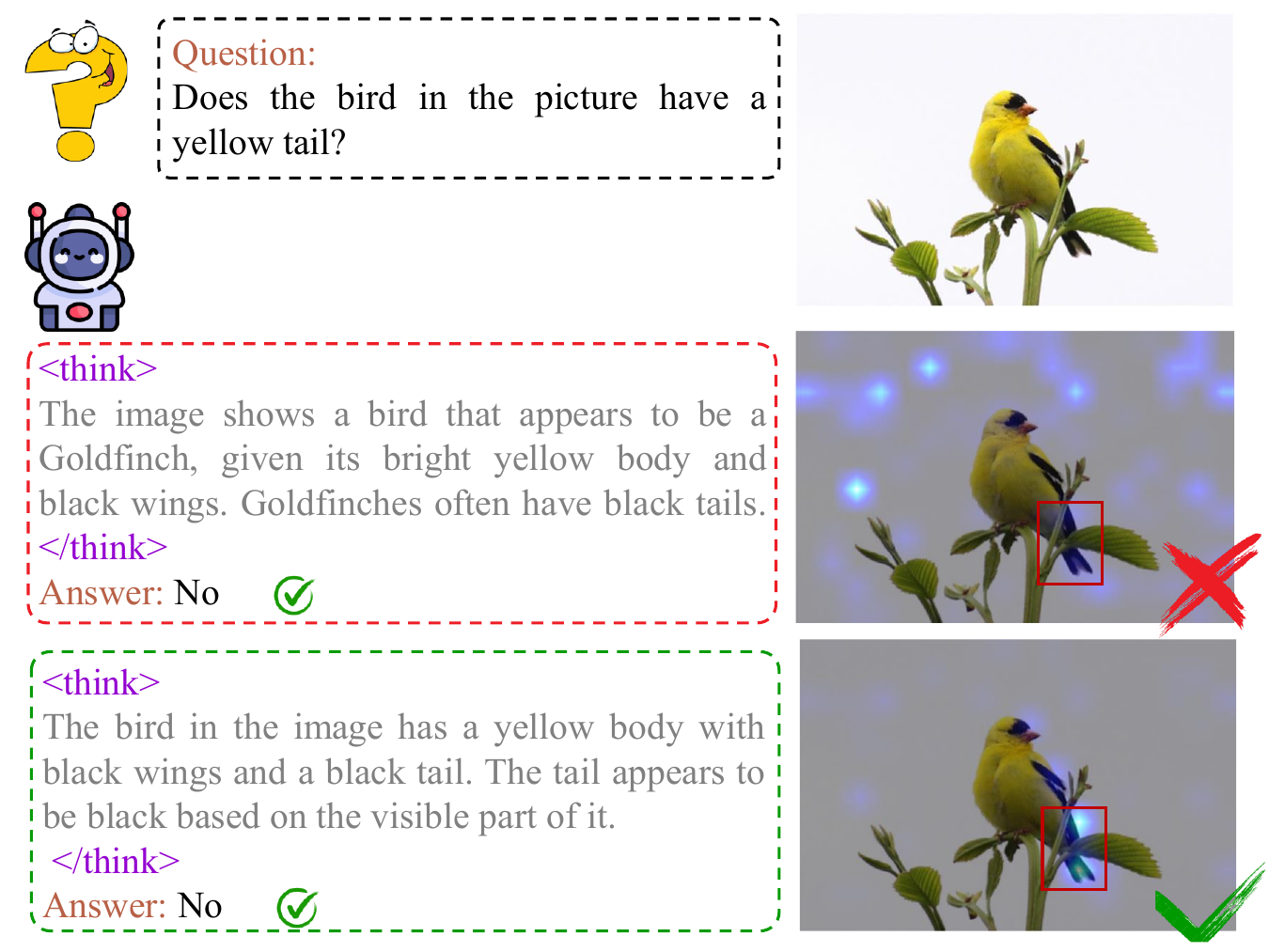}
\caption{\textbf{Main motivation of this work.} Different thinking processes might focus on distinct regions of an image, even if they arrive at the correct answer. Unfaithful thinking processes either focus on irrelevant parts of the image or fail to consider the image.}
\label{teaser}
\vspace{-1.0em}
\end{figure}

In this work, we aim to enhance the interpretability and faithfulness of VLM reasoning. We introduce a saliency-map method that visualizes how visual information influences generated tokens. Specifically, we decompose token logits into first-order contributions~\citep{elhage2021mathematical} from all context tokens and render the contributions from visual tokens as saliency maps. These maps highlight image regions most relevant to each generated token and require no extra forward or backward passes, making the approach computationally efficient. We further extend the framework to trace how visual tokens propagate to answer tokens through the thinking tokens as a bottleneck, revealing image regions that drive both the reasoning process and the final answer. As illustrated in Fig.~\ref{teaser}, different reasoning traces pull from different parts of the image: some emphasize regions truly pertinent to the question, while others attend to irrelevant objects or background. We regard the latter as a potential misuse of visual information or a form of hallucination.

Building on these insights, we propose \textbf{Saliency-R1}, a novel framework that aligns VLM saliency maps with human annotations. We curate a VQA dataset with bounding boxes marking image regions essential for answering each question. For each generated response, we compute the ratio of saliency mass inside the annotated boxes to the saliency mass over the whole image and use this ratio as a reward signal in Group Relative Policy Optimization (GRPO)~\citep{shao2024deepseekmath}. This training process encourages the model to concentrate on task-relevant regions. Extensive experiments across diverse benchmarks show that our approach not only steers reasoning towards critical visual areas but also enhances perceptual grounding and overall performance. Our main contribution can be summarized as:
\begin{itemize}
\item We build a novel attention-based saliency-map technique, which highlights the token-wise important regions efficiently without additional forward or backward passes.
\item We propose an attention rollout mechanism that traces how visual tokens contribute to answer tokens, using the thinking tokens as a bottleneck for visualization.
\item We use the saliency–annotation alignment score as a reward and apply GRPO post-training to enhance the faithfulness and interpretability of VLM reasoning.
\end{itemize}

\section{Related Work}
\noindent \textbf{Explanation of Vision-language Models.}
Our paper mainly focus on the visual explanations of VLMs via the activation maps for the input images. While traditional saliency maps techniques such as Integrated Gradient~\citep{sundararajan2017axiomatic}, CAM~\citep{zhou2016learning}, Grad-CAM,~\citep{selvaraju2017grad}, Rollout~\citep{abnar2020quantifying}, LRP~\citep{lapuschkin2019unmasking}, LIME~\citep{ribeiro2016should}, and SHAP~\citep{NIPS2017_7062} have been applied to interpret the VLMs~\citep{enguehard2023sequential,zhang2024redundancy,ben2024lvlm,gong2025boosting,zimmermann2025needle}, they were built for conventional single-output models (e.g., CNNs, ViTs) and remain suboptimal for sequential prediction. There are also works~\citep{li2025token,shen2025glimpse} design special saliency map techniques for VLMs. However, most of these methods are computationally intensive, as they rely on calculating gradients or perturbing the input and performing multiple forward passes. In contrast, our approach eliminates the need for additional forward or backward passes, enabling real-time computation and seamless integration into the training pipeline.

\noindent \textbf{Vision-language Model Reasoning.} 
In parallel with advances in LLM reasoning capabilities~\citep{openai_learning_to_reason,guo2025deepseek}, a growing body of work targets improving the reasoning of VLMs~\citep{luan2024textcot,mitra2024compositional,zhang2024improve} via either supervised fine-tuning (SFT)~\citep{shao2024visual,sun2025enhancing} or reinforcement learning (RL)~\citep{yao2024mulberry,meng2025mm,thawakar2025llamav,wu2025mmsearch,yang2025r1,xia2025mirage}. Beyond standard LLM reasoning techniques, many VLM studies adopt a ``thinking with images'' paradigm~\citep{su2025thinking}, leveraging specialized visual tools~\citep{liu2025seg,liu2025visionreasoner,zhang2025chain} or image-generation modules~\citep{jiang2025t2i,deng2025emerging,duan2025got} to make intermediate reasoning steps visible. A common strategy is to generate zoomed-in intermediate views that help the model focus on salient regions. In contrast to such explicit zoom-in methods, our approach strengthens a VLM’s intrinsic ability to attend to the most informative areas of an image, eliminating the need for external visualization tools. In addition, our method can also serve as a complement to ``thinking with images'' paradigm by integrating the use of both approaches.

\noindent \textbf{Attention Manipulation in Vision-language Models.}
Our motivation for introducing saliency-map alignment rewards stems from prior findings that steering attention can enhance LLM and VLM performance. LLMs often display biased attention over long contexts, for example, underutilizing middle segments~\citep{liu2023lost} or over-focusing on initial tokens~\citep{xiao2023efficient}. To address this, works have explored fine-tuning~\citep{an2024make}, input-adaptive calibration~\citep{yu2024unveiling}, and position-specific interventions~\citep{yu2024mitigate}. For more targeted control, PASTA~\citep{zhang2024tell} reweights attention toward user-selected spans, while INSTABOOST~\citep{guardieiro2025instruction} and SpotLight~\citep{venkateswaran2025spotlight} emphasize instruction tokens to improve instruction following. In VLMs, ADAPTVIS~\citep{chen2025why} adaptively sharpens or smooths visual attention based on response confidence. To our knowledge, we are the first to align visual attention with human annotations during the post-training phase, thereby bypassing the necessity of precise steering for inference.
\section{Method}
\begin{figure*}[t]
\centering
\includegraphics[width=\textwidth]{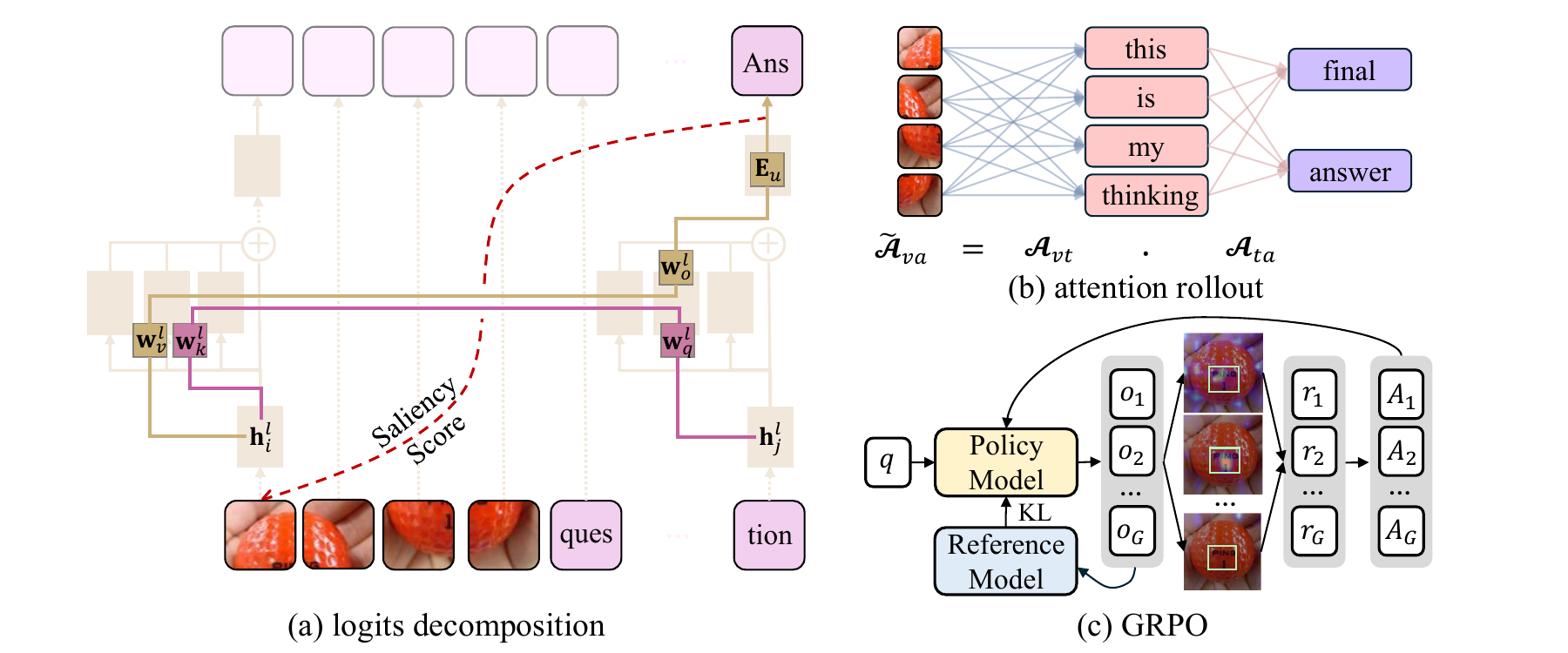}
\caption{\textbf{Overview of our method.} (a) Illustration of saliency map techniques based on logits decomposition. (b) Illustration of attention rollout for generating saliency maps with thinking tokens as the bottleneck. (c) GRPO with saliency maps alignment reward.}
\label{framework}
\end{figure*}

The Saliency-R1 framework is designed to encourage the model to produce a more faithful vision-language reasoning. Faithfulness is evaluated by measuring the alignment between the saliency maps generated during reasoning and the annotated ground truth. To do so, it is crucial to design a saliency-map technique that is both faithful and computationally efficient (Sec.~\ref{sec:3.1}). Moreover, the saliency maps must reflect the entire reasoning trajectory from the thinking process to the final answer (Sec.~\ref{sec:3.2}). Lastly, we should find an appropriate optimization algorithm to effectively enforce this alignment (Sec.~\ref{sec:3.3}). We provide an overview of the proposed framework in Fig.~\ref{framework}.

\subsection{Token-level Saliency-map Generation via Logits Decomposition}
\label{sec:3.1}
As most existing VLMs are based on the transformer architecture~\citep{vaswani2017attention}, we first provide a brief review of the transformer structure. Transformer is a residual network composed of $L$ layers, where each layer consists of a Multi-Head Self-Attention (MSA) mechanism followed by an FFN block. The input texts are first tokenized and then embedded into token embeddings. The images are also encoded into visual tokens with a visual encoder, usually a ViT~\citep{dosovitskiy2020image}. Typical VLMs adopt a decoder-only paradigm, using input tokens along with previously generated tokens to predict the next tokens. At timestep $i$, we denote all past tokens as $\textbf{X}_i=[t_1,t_2,\cdots, t_i]$. With the residual connections, the output of layer $l$ can be expressed as:
\begin{equation*}
\textbf{h}_i^l = \textbf{h}_i^{l-1}+\textbf{A}_i^l + \textbf{F}_i^l,
\end{equation*}
where $\textbf{A}_i^l = \text{ATTN}^l(\textbf{h}_1^{l-1}, \cdots, \textbf{h}_i^{l-1})$ represents the output of the MSA block, and $\textbf{F}_i^l = \text{FFN}^l(\textbf{A}_i^l + \textbf{h}_i^{l-1})$ represents the output of the FFN block\footnote{For simplicity, we ignore the layer normalization layer in the derivation. We will explain it in the Appendix.}. The $L$-th layer output is used to compute the final probability distribution $y$ by multiplying the unembedded matrix $\textbf{E}u\in\mathbb{R}^{d\times N}$. Moreover, the output of the attention layer $\textbf{A}_i^l$ can be represented as a sum of head outputs, each being a weighted sum of value-output vectors on all positions:
\begin{equation*}
\begin{split}
\textbf{A}_i^l&=\sum_{j=1}^H\sum_{p=1}^i\alpha_{i,j,p}^l{\mathbf{W}^l_{o{\color{black},j}}}{{\mathbf{W}^l_{v{\color{black},j}}}}\textbf{h}_p^{l-1}, \\
\alpha^l_{i,j,p}&=\text{softmax}(\mathbf{W}^l_{q,j}\textbf{h}_i^{l-1}\cdot \mathbf{W}^l_{k,j}\textbf{h}_p^{l-1}),
\end{split}
\end{equation*}
where $H$ is the number of attention heads, ${\mathbf{W}_{o{\color{black},j}}^l}\in\mathbb{R}^{d\times d/H}$, $\mathbf{W}_{k,j}^l\in\mathbb{R}^{d/H\times d}$, $\mathbf{W}_{q,j}^l\in\mathbb{R}^{d/H\times d}$, ${\mathbf{W}_{v{\color{black},j}}^l}\in\mathbb{R}^{d/H\times d}$ are the output, key, query, and value matrices. As a result, the final output can be decomposed as:
\begin{equation*}
\textbf{h}_i^L = \textbf{h}_i^0 + \sum_{l=1}^L\textbf{F}_i^l+\sum_{l=1}^L\sum_{j=1}^H\sum_{p=1}^i\alpha_{i,j,p}^l{\mathbf{W}^l_{o,j}}{{\mathbf{W}^l_{v,j}}}\textbf{h}_p^{l-1}.
\end{equation*}
While it is difficult to attribute the first two terms to specific context tokens, the third term consists of the direct contributions of individual context tokens. As a result, we can explicitly trace each token's direct impact on the final logits (Fig.~\ref{framework} (a)). Specifically, the total logits for predicting $t_{i+1}$ are $\textbf{h}_i^L\textbf{E}_u$, where the direct contribution of the token $t_p$ is 
\begin{equation}
\label{eq:3}
c_p=\sum_{l=1}^L\sum_{j=1}^H\alpha_{i,j,p}^l{\mathbf{W}^l_{o{\color{black},j}}}{{\mathbf{W}^l_{v{\color{black},j}}}}\textbf{h}_p^{l-1}\textbf{E}_u.
\end{equation}
We can select the logits contribution that corresponds to the visual tokens, and rearrange according to the orders of the visual patches. We also apply a $ReLU$ function to ignore all the negative contributions. This gives saliency maps highlighting the important regions for the token prediction.  

The way we build the saliency maps simplifies the problem by only considering the direct contribution of tokens. The generated saliency maps may not be the most faithful ones, compared with gradient-based methods or perturbation-based methods. However, as our ultimate goal is to use the saliency maps to guide the training, the efficiency is also very important. Our proposal does not requires additional backward process or multiple forward process. Many attention implementations naturally support returning attention weights, and the term ${{\mathbf{W}^l_{v,j}}}\textbf{h}_p^{l-1}$ is also available within KV cache~\citep{shi2024keep}. Therefore, the saliency maps do not incur extra computation. In addition, previous literature pointed out that the indirect contribution is minor compared to the direct contribution~\citep{elhage2021mathematical,yu2023neuron,gandelsman2024interpreting}. Therefore, aligning the dominant contribution is sufficient to steer the model’s attention to the correct regions.

\begin{table*}[tp]
\centering
\renewcommand\arraystretch{1}
\captionof{table}{\textbf{Faithfulness experiment results.} We use Qwen2.5-VL-3B-Instruct for experiments. Our saliency map technique achieves comparable or better faithfulness to SOTA methods regarding the deletion and insertion metrics. We leave the results on GranDf dataset in the Appendix. The best metric is \textbf{bold} and the second best metric is \underline{underlined}.}
\label{faithfulness}
\resizebox{1\textwidth}{!}{
\begin{tabular}{lcccccccccccc}
\toprule
& \multicolumn{6}{c}{\textbf{COCO Caption}}& 
\multicolumn{6}{c}{\textbf{OpenPSG}}\\
\cmidrule(lr){2-7} \cmidrule(lr){8-13}
\textbf{Method}&\multicolumn{3}{c}{\textbf{Deletion} $\downarrow$}&\multicolumn{3}{c}{\textbf{Insertion}  $\uparrow$}&\multicolumn{3}{c}{\textbf{Deletion} $\downarrow$}&\multicolumn{3}{c}{\textbf{Insertion}  $\uparrow$}\\
\cmidrule(lr){2-4} \cmidrule(lr){5-7} \cmidrule(lr){8-10} \cmidrule(lr){11-13}
&\textbf{5\%}&\textbf{15\%}&\textbf{30\%}&\textbf{5\%}&\textbf{15\%}&\textbf{30\%}&{\textbf{5\%}}\%&\textbf{15\%}&\textbf{30\%}&\textbf{5\%}&\textbf{15\%}&\textbf{30\%}\\
\midrule
CAM~\citep{zhou2016learning}&86.19&82.44&78.34&-0.10&2.47&28.02&87.29&82.50&76.56&-7.12&-2.33&16.10\\
Grad-CAM~\citep{selvaraju2017grad}&86.19&82.44&78.34&-0.10&2.47&28.02&87.29&82.50&76.56&-7.12&-2.33&16.10\\
Raw Attention&84.44&78.90&73.50&-0.12&-0.08&15.95&85.07&77.42&68.68&-7.13&-6.46&10.56\\
Attention Rollout~\citep{abnar2020quantifying}&84.75&71.57&58.48&-0.11&0.03&38.43&87.84&73.55&\underline{59.33}&-8.38&-2.75&24.36\\
ATTN-LRP~\citep{achtibat2024attnlrp}&\underline{76.42}&\underline{64.22}&\underline{52.92}&-0.06&12.10&\textbf{45.67}&84.55&\underline{71.45}&58.50&-11.77&3.93&\textbf{39.66}\\
TAM~\citep{li2025token}&83.91&79.33&73.29&\underline{0.02}&\textbf{23.54}&\underline{45.24}&\underline{82.90}&77.03&69.82&\underline{-3.69}&\underline{8.40}&30.92\\
\rowcolor{backcolour}\textbf{Ours}&\textbf{70.96}&\textbf{59.45}&\textbf{50.34}&\textbf{0.06}&\underline{14.39}&45.22&\textbf{71.52}&\textbf{59.31}&\textbf{49.14}&\textbf{-2.71}&\textbf{9.18}&\underline{35.87}\\
\bottomrule
\end{tabular}}
\end{table*}

\subsection{Holistic Saliency-map Aggregation with Thinking Tokens as Bottleneck}
\label{sec:3.2}
A typical reasoning model first produces a chain of thought and then generates the final answers based on this reasoning context. We assume that the thinking process begins by extracting information from the input images and prompts, followed by summarization and reasoning. The final answers then structure the conclusions drawn from this process into the required output format. To ensure faithful reasoning, we aim for the answer generation to depend on the thinking process, while the thinking process itself should effectively focus on the relevant regions of the input images. To use saliency maps to visualize the process, we hope they could reflect how visual information flows to the answer tokens, with the thinking tokens serving as a bottleneck.

We utilize heuristic attention rollout to quantify how the visual information is transferred to answer tokens through thinking tokens (see Fig.~\ref{framework} (b)). At layer $l$ and head $h$, we use $\mathcal{A}^{l,h}_{vt}$ to denote the attention weight matrix between visual tokens and thinking tokens, where the entry $(i,j)$ denotes the attentions weights from the $i$-th visual token to the $j$-th thinking token. Similarly, $\mathcal{A}^{l,h}_{ta}$ denotes the weight matrix between thinking tokens and answer tokens. Therefore, $\tilde{\mathcal{A}}^{l,h}_{va}=\mathcal{A}^{l,h}_{vt}\mathcal{A}^{l,h}_{ta}$ measures the attention weight from visual tokens to the answer tokens, with thinking tokens as the bottleneck. For a specific answer token, we can utilize Eq.~\ref{eq:3} to visualize the transitional contribution of visual tokens, by replacing the attention weights with the elements in the corresponding column of $\tilde{\mathcal{A}}^{l,h}_{va}$. 

Finally, we form the holistic saliency map by directly summing the per-token maps. We do not normalize the columns of $\mathcal{A}^{l,h}_{vt}$, $\mathcal{A}^{l,h}_{ta}$ or $\tilde{\mathcal{A}}^{l,h}_{va}$ because some answer tokens (e.g., prepositions) receive little contribution from thinking tokens, and similarly some thinking tokens receive little contribution from visual tokens. Such tokens should naturally have a smaller influence on the holistic map. In addition, since the saliency maps reflect logit contributions, low-confidence answer tokens with small logits also contribute less to the holistic saliency map.

\subsection{Boosting Saliency-map Alignment via GRPO}
\label{sec:3.3}
Once we obtain the holistic saliency map, we compare it with human annotation to measure the alignment. Specifically, we draw bounding box on the input image, which highlights the part of the figures important for answering the question. We resize the saliency map to be the same size as the input image, and calculate the recall to represent the alignment score, i.e.,
\begin{equation*}
\text{Alignment\_Score} = \frac{\sum_{i \in \text{ Bounding Box}} \text{Saliency\_Score}(i)}{\sum_{i \in \text{Image}} \text{Saliency\_Score}(i)}.
\end{equation*}

A faithful vision-language reasoning process should achieve a high alignment score, as it actually look at the critical regions of the image relevant to the task. In contrast, hallucinated responses often disregard the image entirely or fixate on unrelated background elements. Likewise, unfaithful chain-of-thoughts rely on shortcuts from visual tokens directly to answer tokens, with minimal information flowing through the thinking bottleneck, leading to low alignment scores of the transitional saliency map. To enforce higher alignment score, we use GRPO~\citep{shao2024deepseekmath} as the optimization algorithm (see Fig.~\ref{framework} (c)). During the training process, GRPO samples a group of outputs based on the given question from the old policy $\pi_{\theta_{\text{old}}}$ and then optimizes the policy model $\pi_\theta$ by maximizing the following objective:
\begin{equation*}
{\small
\begin{split}
&\mathcal{J}_{\text{GRPO}}(\theta)= \mathbb{E}_{[q\sim\mathcal{D}, \{o_i\}_{i=1}^G\sim\pi_{\theta_{\text{old}}}(\cdot|q)]}\frac{1}{G}\sum_{i=1}^G\left(M_i-\beta\mathbb{D}_{KL}(\pi_\theta\Vert\pi_{\text{ref}})\right),\\
&M_i = \min\left(\frac{\pi_{\theta}(o_i|q)}{\pi_{\theta_{\text{old}}}(o_i|q)}A_i, \operatorname{clip}\left(\frac{\pi_{\theta}(o_i|q)}{\pi_{\theta_{\text{old}}}(o_i|q)},1-\epsilon, 1+\epsilon\right)A_i\right), \\
&\mathbb{D}_{KL}(\pi_\theta\Vert\pi_{\text{ref}})=\frac{\pi_\text{ref}(o_i|q)}{\pi_\theta(o_i|q)}-\log\frac{\pi_{\text{ref}}(o_i|q)}{\pi_\theta(o_i|q)}-1,
\end{split}}
\end{equation*}
where $q$ represents the input question drawn from the dataset $\mathcal{D}$, and $\{o_1,o_2,\cdots,o_G\}$ denotes the generated text response. $\mathbb{D}_{KL}$ is the KL-divergence measure, while $\epsilon$ and $\beta$ are hyper-parameters. $A_i$ indicates the advantage, computed using a group of rewards $\{r_1, r_2, \cdots, r_G\}$ corresponding to the outputs within each group. The standardized rewards are used as the advantage.

Different reward functions can lead the model toward different optimization directions and final performance outputs. The reward function in our Saliency-R1 framework consists of three components:
\begin{equation*}
\mathcal{R}_{\text{overall}} = \mathcal{R}_{\text{accuracy}}+ \mathcal{R}_{\text{format}} + \mathcal{R}_{\text{saliency}}.
\end{equation*}

$\mathcal{R}_{\text{format}}$ enforces the model to follow the instruction. The model receives a reward of 1 if the reasoning process is enclosed in ``\textless think\textgreater\textless/think\textgreater'' tags and the final answer is followed afterwards. Otherwise, the reward score is 0. The accuracy reward $\mathcal{R}_{\text{accuracy}}$ is calculated using LLM-as-judge~\citep{yang2025visionthink}, where 0 indicates an incorrect answer and 1 indicates a correct answer. We further use the alignment score as the saliency reward $\mathcal{R}_{\text{saliency}}$. To this end, responses with higher saliency alignment are treated as exhibiting better reasoning. The training process guides the model towards the direction of generating more responses with high alignment scores, thereby being more interpretable.

\section{Experiments}

\subsection{Experimental Setup}
\textbf{Training Data.} Following standard practice~\citep{guo2025deepseek}, we begin with SFT for a cold start. We build our SFT dataset by filtering Vision-R1-cold~\citep{huang2025vision} to remove multi-round conversations and extremely long responses, yielding 272,881 samples. For GRPO, we construct the saliency-r1-8k dataset. It is composed of 8,080 samples, down-sampled from the Visual-CoT dataset~\citep{shao2024visual}, which consists of VQA samples for OCR, fine-grained understanding, general VQA, charts understanding, and relation reasoning. Each sample includes an image, a question, a ground-truth answer, and a bounding box marking the image regions relevant to the question. Further dataset statistics and curation details are provided in the Appendix.

\noindent \textbf{Implementation Details.} Our main experiments utilize Qwen2.5-VL~\citep{bai2025qwen2} (3B and 7B) as the base model. SFT is trained using llama-factory~\citep{zheng2024llamafactory} and GRPO is trained with the TRL framework~\citep{vonwerra2022trl}. We use a total batch size of 64 with mixed precision. The KL divergence coefficient is 0.001, and we sample 8 rollouts per step. The policy model is optimized with LoRA~\citep{hu2022lora} (rank 16) using an initial learning rate of $10^{-5}$. GPT-4o-mini~\cite{hurst2024gpt} serves as the LLM judge to provide accuracy rewards. All training runs are executed on 8 NVIDIA A6000 GPUs.

\subsection{Evaluation of Saliency Maps Faithfulness}
We begin by evaluating the faithfulness of our saliency methods using deletion and insertion tests. In deletion, we progressively remove the most salient image pixels in descending order; in insertion, we add those pixels back onto a blank canvas. To enable fair comparison, the model’s mean self-log-likelihood is linearly normalized using the blank image and the original, unperturbed output as references. We report normalized log-likelihood at 5\%, 15\%, and 30\% pixel perturbation. Lower deletion scores and higher insertion scores indicate that the highlighted regions have a stronger causal effect on the model’s responses.

We conduct experiments using Qwen2.5-VL-3B as the VLM, and compare our method with several baselines, including CAM~\citep{zhou2016learning}, Grad-CAM~\citep{selvaraju2017grad}, raw attention, attention rollout~\citep{abnar2020quantifying}, ATTN-LRP~\citep{achtibat2024attnlrp}, and TAM~\citep{li2025token}. The evaluation is performed on three datasets, namely the COCO Caption dataset~\citep{chen2015microsoft}, GranDf~\citep{rasheed2024glamm}, and OpenPSG~\citep{zhou2024openpsg}. For each dataset, we sample 1k images for evaluation. The results are shown in Table~\ref{faithfulness}. Our method attains faithfulness on par with, or better than, the baselines. For example, on COCO Captions, the deletion score is lower than the second-best approach by 5.46\%, 4.77\%, and 2.57\% at 5\%, 15\%, and 30\% pixel perturbation, respectively. This indicates that, even though we focus solely on direct contributions, the method still faithfully captures the relative importance of visual patches. In addition, our method is very efficient. Unlike baselines such as ATTN-LRP (which requires backpropagation) and TAM (which solves an optimization problem), our approach relies only on simple matrix operations, making it especially suitable for integration into post-training pipelines.

\begin{table*}[t]
  \centering
  \caption{\textbf{Effective Performance Compared to the SOTA Model.} Our models are based on Qwen2.5-VL-7B-Instruct and Qwen2.5-VL-BB-Instruct. The reported performance of the base models are evaluated by lmms-eval~\citep{zhang2024lmms}.}
  \begin{adjustbox}{width=\textwidth}
  \renewcommand{\arraystretch}{1.5}
  \setlength{\tabcolsep}{2pt}
  \begin{tabular}{@{}lcccccccccc@{}}
    \toprule
    \multirow{2}{*}{\textbf{Method}} & \textbf{MMMU-Pro} & \textbf{MMBench} & \textbf{POPE} & \textbf{MME} & \textbf{MME-RW
} & \textbf{MMStar}&\textbf{ChartQA}&\textbf{IllusionVQA}&\textbf{ScienceQA}&\textbf{SalBench}\\ 
    \cline{2-11}
     & test & en\_test & test & test & test & test & test & loc\_test & img\_test & p3\_test\\
    \midrule
    \rowcolor{backcolour}
    \multicolumn{11}{l}{\textit{Closed-source Models}} \\
    GPT-4o~\citep{hurst2024gpt} & 54.0 & 83.4 & 85.6 & 2329 & 45.2 & 63.9 & 85.7 & 49.1 & 90.1 & 89.2 \\
    Claude-3.5 Sonnet~\citep{anthropic2024claude35} & 55.0 & 82.6 & - & 1920 & 51.6 & 62.2 & 90.8 & 47.4 & 88.9 & 86.7 \\
    Gemini-1.5-Pro~\citep{team2024gemini} & 49.4&73.9&88.2&-&38.2&59.1&-&47.3 & 85.7 & - \\
    \hdashline
    \rowcolor{backcolour} \multicolumn{11}{l}{\textit{Open-source General Models}} \\
    Cambrain-1-8B~\citep{tong2024cambrian} & - & 75.9 & 86.4 & 1803 & 42.7 & - & 72.6 &- & 81.0 & - \\ 
    InternVL2-8B~\citep{chen2024expanding} & 32.5 & 81.7 & 84.2 & 2210 & 53.5 & 61.5 & - & 28.3 & 97.1 & - \\
    InternVL2.5-8B~\citep{chen2024expanding} & 38.2 & 84.6 & 90.6 & 2344 & 58.1 & 63.2 & - &- & 98.4 & - \\
    LLaVA-OneVision-7B~\citep{li2024llava} & -& - & 88.4 & 1998 & 57.4 & 61.9 & 80.9 & 37.3 & 95.4 & - \\
    MiniCPM-Llama-V-2.5-8B~\citep{yao2024minicpm} &19.6 & 77.2 & 86.7 & 2025 & 45.6 & 51.8 & - & - & 89.2 & - \\
    MiniCPM-V-2.6-8B~\citep{yao2024minicpm} & 27.2 & 78.0 & 83.2 & 2348 & - & 57.5 & 82.4 & - & 96.7 & -\\
    IXC-2.5~\citep{zhang2024internlm} & - & 82.2 & 87.9 & 2229 & 50.0 & 59.9 & 82.2 &-& 96.6 & - \\
    \hdashline
   \rowcolor{backcolour} \multicolumn{11}{l}{\textit{Reasoning Models}} \\
    LLaVA-CoT-11B~\citep{xu2024llava} & -&75.0&-&-&-&54.9&-&-& -& - \\
    LLaVA-Reasoner-8B~\citep{zhang2024improve} & - & - & - & - & - & - & 83.0 & - & - & - \\
    Insight-V-8B~\citep{dong2025insight} & 24.9 & 82.3 & - & 2312 & - & 61.5 & 81.5 & - & - & -\\
    Mulberry-7B~\citep{yao2024mulberry} & - & - & - & 2396 & - & 61.3 & 83.9 & - & - & -\\
    Vision-R1-LlamaV-CI-11B~\citep{huang2025vision} & - & - & - & 2190 & - & 61.4 & 83.9 & - & - & -\\
    \midrule
    \rowcolor{backcolour}
    \multicolumn{11}{l}{\textit{\textbf{Saliency-R1}}} \\
    Qwen2.5-VL-3B~\citep{bai2025qwen2} & 30.8 & 77.8 & 87.5 & 2206 & 52.0 & 56.1 & 83.6 & 35.7 & 81.2 & 40.4 \\
    \qquad+ SFT & 29.9 & 75.8 & 87.1 & 2351 & 56.0 & 55.7 & 84.0 & 38.1 & 90.5 & 51.7 \\
    \qquad+ Saliency-R1 & 31.1 & 76.7 & 87.5 & 2235 & 57.7 & 56.0 & 84.2 & 39.9 & 91.2 & 59.6 \\
    \hdashline
    Qwen2.5-VL-7B~\citep{bai2025qwen2} & 36.2 & 82.8 & 86.7 & 2302 & 58.7 & 62.4 & 84.0 & 37.5 & 88.2 & 49.1 \\
    \qquad+ SFT & 35.5 & 78.8 & 87.6 & 2372 & 60.6 & 60.1 & 87.2 & 33.6 & 93.8 & 57.1 \\
    \qquad+ Saliency-R1 & 37.6 & 81.8 & 88.1 & 2385 & 62.9 & 62.6 & 88.2 & 38.4 & 94.3 & 63.7 \\
    \bottomrule
  \end{tabular}
  \end{adjustbox}
  \label{table:generalqa_result}
\end{table*}

\subsection{Evaluation of Saliency-R1 on General VQA Benchmarks}
We then evaluate the performance of Saliency-R1 on several general VQA benchmarks, mainly focusing on vision-centric tasks, including MMMU-Pro~\citep{yue2024mmmu-pro}, MMBench~\citep{liu2024mmbench},  POPE~\citep{li2023evaluating}, MME~\citep{chaoyou2023mme}, MME-RealWorld (MME-RW)~\citep{zhang2024mme}, MMStar~\citep{chen2024we}, ChartQA~\citep{masry2022chartqa}, IllusionVQA~\citep{shahgir2024illusionvqa}, ScienceQA~\citep{saikh2022scienceqa}, and SalBench~\citep{dahou2025vision}. The evaluation utilizes the lmms-eval framework~\citep{zhang2024lmms}.

\begin{figure*}[t]
\centering
\includegraphics[width=\textwidth]{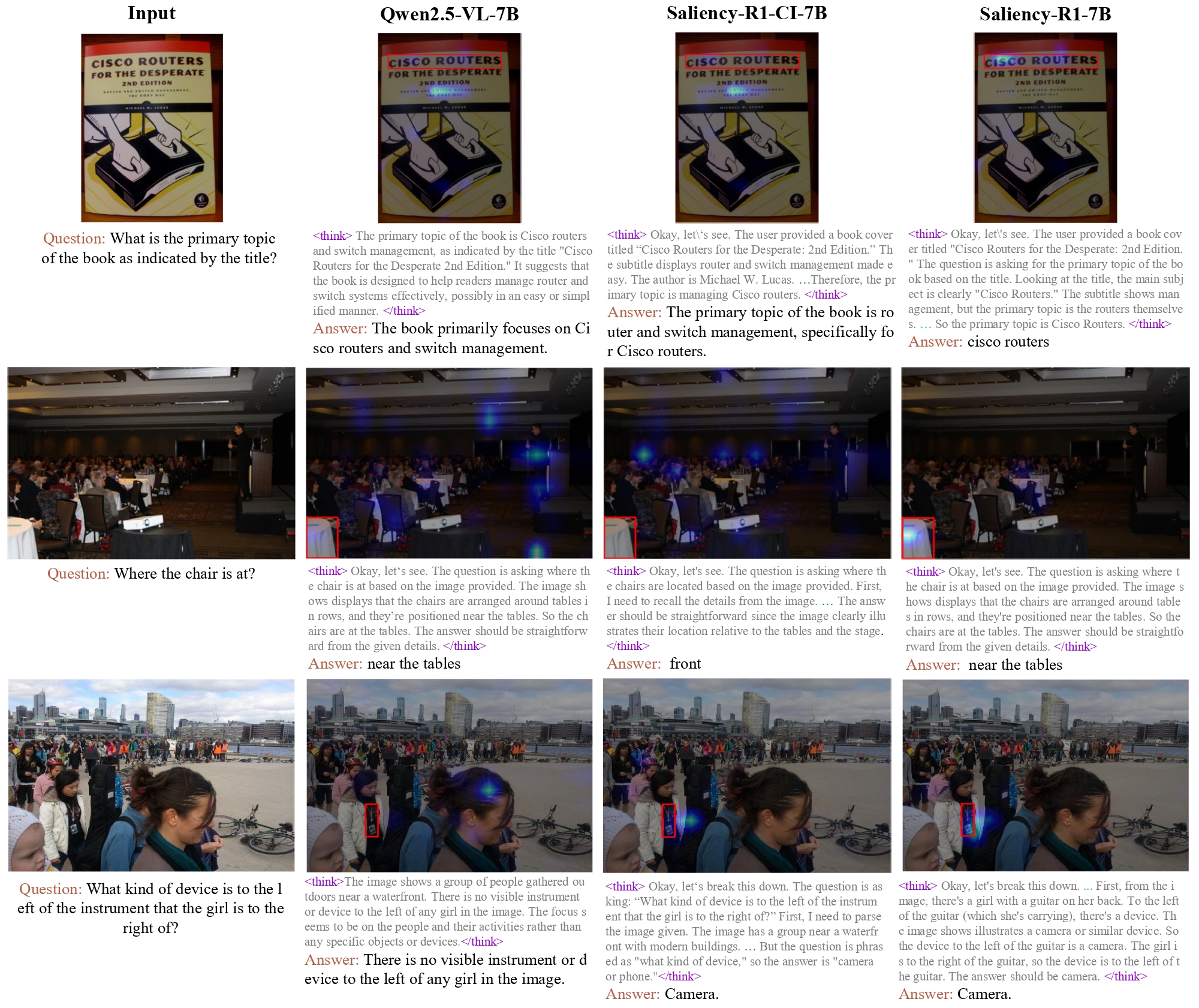}
\caption{\textbf{Qualitative evaluation of interpretability.} We present example responses and their corresponding saliency maps from the base model, the SFT-tuned model (Saliency-R1-CI), and saliency-R1. The ground-truth bounding box is highlighted in {\color{red}red}. Due to space constraints, some nonessential parts of the model responses are omitted; the full versions are provided in the Appendix.}
\label{example}
\end{figure*}

We first compare our saliency-R1 with several baselines. The comparisons include the base models (both Qwen2.5-VL-7B and Qwen2.5-VL-3B) and the cold-started models after SFT without RL (Saliency-R1-CI).  The results are presented in Table~\ref{table:generalqa_result}. We notice that Saliency-R1 can significantly improve the model performance on VQA tasks, especially those vision-centric benchmarks, such as hallucination and OCR benchmarks. For example, Saliency-R1-7B outperforms the base model by 1.4\% and 4.2\% on POPE and ChartQA, respectively. It also improves the model performance on more general VQA tasks. In addition, we observe the improvements are consistent on both 3B and 7B models, showing the universality of Salieny-R1. While SFT alone also improves the reasoning capability of the model, the improvements are minor compared with Saliency-R1. 

We further conduct comparison with several SOTA models, including several closed-source model (i.e., GPT-4o~\citep{hurst2024gpt}, Claude-3.5 Sonnet~\citep{anthropic2024claude35}, and Gemini-1.5-Pro~\citep{team2024gemini}), open-source model (i.e., Cambrian-1-8B~\citep{tong2024cambrian}, InternVL2-8B, InternVL2.5-8B~\citep{chen2024expanding}, LLaVA-OneVision-7B~\citep{li2024llava}, MiniCPM-Llama-V-2.5-8B, MiniCPM-V-2.6-8B~\citep{yao2024minicpm}, and IXC-2.5~\citep{zhang2024internlm}), and large reasoning models (i.e., LLaVA-CoT-11B~\citep{xu2024llava}, LLaVA-Reasoner-8B~\citep{zhang2024improve}, Insight-V-8B~\citep{dong2025insight}, Mulberry-7B~\citep{yao2024mulberry}, and Vision-R1-LlamaV-CI-11B~\citep{huang2025vision}). We notice Saliency-R1 achieves SOTA performance compared with both general models and reasoning models. It even achieves superior performance compared with commercial models on several benchmarks such as MME, MME-RealWorld, and ScienceQA.

\subsection{Evaluation of Interpretability of Saliency-R1}
We further evaluate the interpretability of Saliency-R1. We first give quantitative evaluation on the test set of the saliency-r1-8k dataset. We conduct the pointing game~\citep{wang2020score} and calculate two metrics: the PG metric assigns a hit score if the point with the highest value on the saliency map falls within the object region, defined by the bounding box. The PG accuracy is then calculated by averaging all sample scores; energy-PG measures the ratio of saliency map energy within the bounding box to the entire map, providing a better reflection of the saliency map’s spread. The results are presented in Table~\ref{tab:interpretability}. Saliency-R1 achieves significantly better interpretability. Compared with the base model, the energy-PG and PG are increased by 10.82\% and 14.14\%, respectively. We notice that SFT can also improve the interpretability of the base model by enforcing the model to think following the human-labeled reasoning path. However, it is still inferior to Saliency-R1, which incorporates the saliency alignment constraint explicitly. Combining Saliency-R1 with SFT results yields the best performance.

In Fig.~\ref{example}, we show several sample questions and the corresponding responses. We present the thinking process and the answers generated by Qwen2.5-VL-7B, Saliency-R1-CI, and Saliency-R1. Moreover, we visualize the generated saliency maps and also present the ground truth bounding box for reference. The base model often assigns attention to irrelevant areas, which may lead to wrong answers or correct answers but suboptimal reasoning processes. We further observe that the length of the reasoning given by Saliency-R1-CI is unnecessarily long, with a lot of repetitive and irrelevant content. This may dilute the attention on the critical regions. On the other hand, Saliency-R1 enforces concentrated attention on the relevant regions crucial for answering the question, resulting in more aligned saliency maps and concise thinking processes.

\subsection{Ablation Studies}
\begin{figure}[t]
\centering
\includegraphics[width=0.45\textwidth]{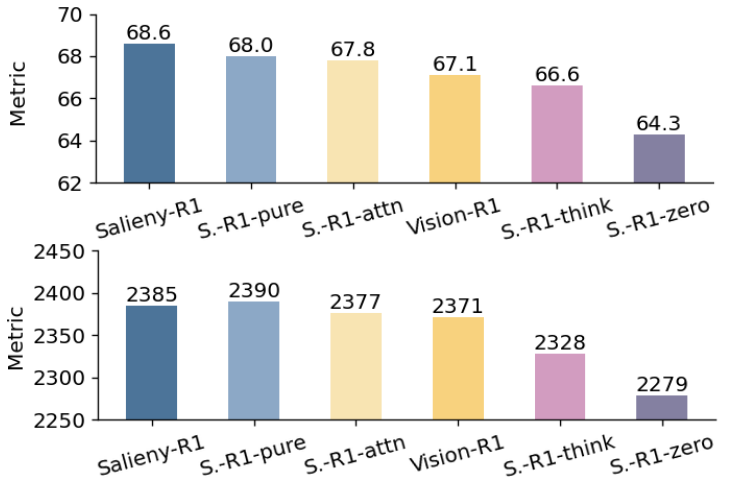}
\caption{\textbf{Ablation Studies.} \textit{Top}: Average metrics on 9 VQA benchmarks. \textit{Bottom}: Metrics on MME.}
\vspace{-3mm}
\label{ablation}
\end{figure}

We conduct extensive ablation studies to examine the effects of individual model design. We use Qwen2.5-VL-7B-Instruct as the base model and evaluate the model performance on the same 10 datasets as in Table~\ref{table:generalqa_result}. The average metrics are reported in Fig.~\ref{ablation}. Because of the large scale difference, we report MME separately from the average over the rest nine datasets. A complete set of results, including additional experiments, is provided in the Appendix.

\begin{table}[tp]
\centering
\renewcommand\arraystretch{1}
\captionof{table}{\textbf{Evaluation of Interpretability.} We calculate the energy-PG and PG metric on the test set of saliency-r1-8k dataset.}
\label{tab:interpretability}
\vspace{-1mm}
\resizebox{0.4\textwidth}{!}{\begin{tabular}{lcc}
\toprule
\textbf{Method}&Energy-PG (\%) $\uparrow$&PG (\%) $\uparrow$\\
\midrule
Qwen2.5-VL-7B~\citep{bai2025qwen2}&19.26&24.49\\
Saliency-R1-CI&23.65&29.04\\
Vision-R1&25.32&32.58\\
Saliency-R1-zero&26.27&32.58\\
\rowcolor{backcolour}Saliency-R1-7B&\textbf{30.08}&\textbf{38.63}\\
\bottomrule
\end{tabular}}
\vspace{-3mm}
\end{table}

\noindent \textbf{Effects of SFT Cold Start.} We evaluate the performance of Saliency-R1 against Saliency-R1-zero, which leverages GRPO to train the model directly without an SFT cold start. The results show that SFT plays a crucial role in significantly enhancing model performance. Given the relatively small size of the base model, its initial reasoning capabilities are limited. The cold start process allows the model to quickly develop fundamental reasoning skills.

\noindent \textbf{Effects of Individual Reward Term.} We then investigate the impact of each reward term. We begin with standard GRPO training without incorporating the saliency alignment reward. This gives the same training protocol as Vision-R1~\cite{huang2025vision}, but using different training data. The results reveal that Saliency-R1 outperforms Vision-R1 by 1.5\% across 9 VQA benchmarks, highlighting the importance of the saliency alignment reward. By aligning saliency maps, the model concentrates on the most critical regions of the image, showing significantly better visual reasoning ability.

We also observed a general positive correlation between saliency alignment and accuracy: when the model attends to the correct region, it tends to produce the correct answer. Moreover, since computing the saliency reward depends on properly formatted outputs, its value already captures formatting correctness. This suggests the saliency reward could be used on its own. To test this hypothesis, we train Saliency-R1-pure using only the saliency alignment reward. The model performs well, with the average accuracy across nine benchmarks just 0.6\% lower than the full Saliency-R1. It even surpasses Saliency-R1 on the MME benchmark. Training with only the saliency reward can lower costs by eliminating the need to evaluate generated answers with another LLM.

\noindent \textbf{Effects of Saliency Maps Techniques.} Finally, we evaluate the impact of our proposed saliency map technique. We replace it with a vanilla attention aggregation and retrain the model, yielding the variant Saliency-R1-attn. This model still outperforms Vision-R1 on general VQA benchmarks, indicating that alignment via attention maps also helps the model ``see'' better. However, its performance lags behind Saliency-R1. We also examine with the version of saliency maps that directly aggregate the thinking tokens' saliency maps (without attention rollout), yielding the version Saliency-R1-think. Still, the results are sub-optimal. Our saliency maps more faithfully capture the reasoning process, aligning them has a more direct and stronger effect on improving the visual ability of the model.

\section{Limitation and Conclusion}
In this work, we propose Saliency-R1, a novel protocol that improves the visual reasoning ability of VLMs via saliency map alignment. Specifically, we propose a real-time saliency-map generation algorithm via logits decomposition, and propose to align the saliency maps with human annotations through GRPO. Evaluation of Saliency-R1 on 10 VQA benchmarks shows its superior performance. 

Our work has certain limitations. Due to the limited computational resources, the training, especially SFT, is conducted on relatively small datasets. As shown in Table 2, some metrics become lower after SFT. This is not desired and can be overcome by collecting more diverse cold-start data. Moreover, we use bounding boxes as the coarse labels to align saliency maps. Finer segmentation labels (e.g., from SAM 3~\cite{carion2025sam}) may further improve performance. Finally, our evaluation is based on relatively small base models. The extension to larger models is left as future work.

\clearpage
\section*{Acknowledgment} This work was supported in part by a grant from the National Natural Science Foundation of China (Project No. 62322318) and in part by a grant from the Research Grants Council of the Hong Kong Special Administrative Region, China (Project No. T45-401/22-N).
{
    \small
    \bibliographystyle{ieeenat_fullname}
    \bibliography{main}

@inproceedings{gong2025concepts,
  title={Concepts from Neurons: Building Interpretable Medical Image Diagnostic Models by Dissecting Opaque Neural Networks},
  author={Gong, Shizhan and Wang, Huayu and Zhang, Xiaofan and Dou, Qi},
  booktitle={International Conference on Information Processing in Medical Imaging},
  pages={3--18},
  year={2025},
  organization={Springer}
}

@inproceedings{cui2024survey,
  title={A survey on multimodal large language models for autonomous driving},
  author={Cui, Can and Ma, Yunsheng and Cao, Xu and Ye, Wenqian and Zhou, Yang and Liang, Kaizhao and Chen, Jintai and Lu, Juanwu and Yang, Zichong and Liao, Kuei-Da and others},
  booktitle={Proceedings of the IEEE/CVF winter conference on applications of computer vision},
  pages={958--979},
  year={2024}
}

@article{lee2025large,
  title={Large language models in finance (finllms)},
  author={Lee, Jean and Stevens, Nicholas and Han, Soyeon Caren},
  journal={Neural Computing and Applications},
  pages={1--15},
  year={2025},
  publisher={Springer}
}

@inproceedings{
parcalabescu2025do,
title={Do Vision \& Language Decoders use Images and Text equally? How Self-consistent are their Explanations?},
author={Letitia Parcalabescu and Anette Frank},
booktitle={The Thirteenth International Conference on Learning Representations},
year={2025},
url={https://openreview.net/forum?id=lCasyP21Bf}
}

@inproceedings{
hao2025can,
title={Can {MLLM}s Reason in Multimodality? {EMMA}: An Enhanced MultiModal ReAsoning Benchmark},
author={Yunzhuo Hao and Jiawei Gu and Huichen Will Wang and Linjie Li and Zhengyuan Yang and Lijuan Wang and Yu Cheng},
booktitle={Forty-second International Conference on Machine Learning},
year={2025},
url={https://openreview.net/forum?id=v26vwjxOEz}
}

@inproceedings{favero2024multi,
  title={Multi-modal hallucination control by visual information grounding},
  author={Favero, Alessandro and Zancato, Luca and Trager, Matthew and Choudhary, Siddharth and Perera, Pramuditha and Achille, Alessandro and Swaminathan, Ashwin and Soatto, Stefano},
  booktitle={Proceedings of the IEEE/CVF Conference on Computer Vision and Pattern Recognition},
  pages={14303--14312},
  year={2024}
}

@article{wei2022chain,
  title={Chain-of-thought prompting elicits reasoning in large language models},
  author={Wei, Jason and Wang, Xuezhi and Schuurmans, Dale and Bosma, Maarten and Xia, Fei and Chi, Ed and Le, Quoc V and Zhou, Denny and others},
  journal={Advances in neural information processing systems},
  volume={35},
  pages={24824--24837},
  year={2022}
}

@article{guo2025deepseek,
  title={Deepseek-r1: Incentivizing reasoning capability in llms via reinforcement learning},
  author={Guo, Daya and Yang, Dejian and Zhang, Haowei and Song, Junxiao and Zhang, Ruoyu and Xu, Runxin and Zhu, Qihao and Ma, Shirong and Wang, Peiyi and Bi, Xiao and others},
  journal={arXiv preprint arXiv:2501.12948},
  year={2025}
}

@inproceedings{parcalabescu2024measuring,
  title={On Measuring Faithfulness or Self-consistency of Natural Language Explanations},
  author={Parcalabescu, Letitia and Frank, Anette},
  booktitle={Proceedings of the 62nd Annual Meeting of the Association for Computational Linguistics (Volume 1: Long Papers)},
  pages={6048--6089},
  year={2024}
}

@article{admoni2025towards,
  title={Towards Large Language Models with Self-Consistent Natural Language Explanations},
  author={Admoni, Sahar and Amir, Ofra and Hallak, Assaf and Ziser, Yftah},
  journal={arXiv preprint arXiv:2506.07523},
  year={2025}
}

@article{elhage2021mathematical,
  title={A mathematical framework for transformer circuits},
  author={Elhage, Nelson and Nanda, Neel and Olsson, Catherine and Henighan, Tom and Joseph, Nicholas and Mann, Ben and Askell, Amanda and Bai, Yuntao and Chen, Anna and Conerly, Tom and others},
  journal={Transformer Circuits Thread},
  volume={1},
  number={1},
  pages={12},
  year={2021}
}

@article{shao2024deepseekmath,
  title={Deepseekmath: Pushing the limits of mathematical reasoning in open language models},
  author={Shao, Zhihong and Wang, Peiyi and Zhu, Qihao and Xu, Runxin and Song, Junxiao and Bi, Xiao and Zhang, Haowei and Zhang, Mingchuan and Li, YK and Wu, Yang and others},
  journal={arXiv preprint arXiv:2402.03300},
  year={2024}
}

@misc{openai_learning_to_reason,
  title = {Learning to Reason with Language Models},
  author = {OpenAI},
  year = {2024},
  url = {https://openai.com/index/learning-to-reason-with-llms/},
  note = {Accessed: 2025-08-28}
}

@article{luan2024textcot,
  title={Textcot: Zoom in for enhanced multimodal text-rich image understanding},
  author={Luan, Bozhi and Feng, Hao and Chen, Hong and Wang, Yonghui and Zhou, Wengang and Li, Houqiang},
  journal={arXiv preprint arXiv:2404.09797},
  year={2024}
}

@inproceedings{mitra2024compositional,
  title={Compositional chain-of-thought prompting for large multimodal models},
  author={Mitra, Chancharik and Huang, Brandon and Darrell, Trevor and Herzig, Roei},
  booktitle={Proceedings of the IEEE/CVF Conference on Computer Vision and Pattern Recognition},
  pages={14420--14431},
  year={2024}
}

@article{zhang2024improve,
  title={Improve vision language model chain-of-thought reasoning},
  author={Zhang, Ruohong and Zhang, Bowen and Li, Yanghao and Zhang, Haotian and Sun, Zhiqing and Gan, Zhe and Yang, Yinfei and Pang, Ruoming and Yang, Yiming},
  journal={arXiv preprint arXiv:2410.16198},
  year={2024}
}

@article{shao2024visual,
  title={Visual cot: Advancing multi-modal language models with a comprehensive dataset and benchmark for chain-of-thought reasoning},
  author={Shao, Hao and Qian, Shengju and Xiao, Han and Song, Guanglu and Zong, Zhuofan and Wang, Letian and Liu, Yu and Li, Hongsheng},
  journal={Advances in Neural Information Processing Systems},
  volume={37},
  pages={8612--8642},
  year={2024}
}

@article{sun2025enhancing,
  title={Enhancing Step-by-Step and Verifiable Medical Reasoning in MLLMs},
  author={Sun, Haoran and Jiang, Yankai and Lou, Wenjie and Zhang, Yujie and Li, Wenjie and Wang, Lilong and Liu, Mianxin and Liu, Lei and Wang, Xiaosong},
  journal={arXiv preprint arXiv:2506.16962},
  year={2025}
}

@article{huang2025vision,
  title={Vision-r1: Incentivizing reasoning capability in multimodal large language models},
  author={Huang, Wenxuan and Jia, Bohan and Zhai, Zijie and Cao, Shaosheng and Ye, Zheyu and Zhao, Fei and Xu, Zhe and Hu, Yao and Lin, Shaohui},
  journal={arXiv preprint arXiv:2503.06749},
  year={2025}
}

@article{meng2025mm,
  title={Mm-eureka: Exploring the frontiers of multimodal reasoning with rule-based reinforcement learning},
  author={Meng, Fanqing and Du, Lingxiao and Liu, Zongkai and Zhou, Zhixiang and Lu, Quanfeng and Fu, Daocheng and Han, Tiancheng and Shi, Botian and Wang, Wenhai and He, Junjun and others},
  journal={arXiv preprint arXiv:2503.07365},
  year={2025}
}

@article{thawakar2025llamav,
  title={Llamav-o1: Rethinking step-by-step visual reasoning in llms},
  author={Thawakar, Omkar and Dissanayake, Dinura and More, Ketan and Thawkar, Ritesh and Heakl, Ahmed and Ahsan, Noor and Li, Yuhao and Zumri, Mohammed and Lahoud, Jean and Anwer, Rao Muhammad and others},
  journal={arXiv preprint arXiv:2501.06186},
  year={2025}
}

@article{wu2025mmsearch,
  title={MMSearch-R1: Incentivizing LMMs to Search},
  author={Wu, Jinming and Deng, Zihao and Li, Wei and Liu, Yiding and You, Bo and Li, Bo and Ma, Zejun and Liu, Ziwei},
  journal={arXiv preprint arXiv:2506.20670},
  year={2025}
}

@article{yang2025r1,
  title={R1-onevision: Advancing generalized multimodal reasoning through cross-modal formalization},
  author={Yang, Yi and He, Xiaoxuan and Pan, Hongkun and Jiang, Xiyan and Deng, Yan and Yang, Xingtao and Lu, Haoyu and Yin, Dacheng and Rao, Fengyun and Zhu, Minfeng and others},
  journal={arXiv preprint arXiv:2503.10615},
  year={2025}
}

@article{yao2024mulberry,
  title={Mulberry: Empowering mllm with o1-like reasoning and reflection via collective monte carlo tree search},
  author={Yao, Huanjin and Huang, Jiaxing and Wu, Wenhao and Zhang, Jingyi and Wang, Yibo and Liu, Shunyu and Wang, Yingjie and Song, Yuxin and Feng, Haocheng and Shen, Li and others},
  journal={arXiv preprint arXiv:2412.18319},
  year={2024}
}

@article{xia2025mirage,
  title={MIRAGE: Towards AI-Generated Image Detection in the Wild},
  author={Xia, Cheng and Lin, Manxi and Tan, Jiexiang and Du, Xiaoxiong and Qiu, Yang and Zheng, Junjun and Kong, Xiangheng and Jiang, Yuning and Zheng, Bo},
  journal={arXiv preprint arXiv:2508.13223},
  year={2025}
}

@article{su2025thinking,
  title={Thinking with images for multimodal reasoning: Foundations, methods, and future frontiers},
  author={Su, Zhaochen and Xia, Peng and Guo, Hangyu and Liu, Zhenhua and Ma, Yan and Qu, Xiaoye and Liu, Jiaqi and Li, Yanshu and Zeng, Kaide and Yang, Zhengyuan and others},
  journal={arXiv preprint arXiv:2506.23918},
  year={2025}
}

@article{liu2025visionreasoner,
  title={VisionReasoner: Unified Visual Perception and Reasoning via Reinforcement Learning},
  author={Liu, Yuqi and Qu, Tianyuan and Zhong, Zhisheng and Peng, Bohao and Liu, Shu and Yu, Bei and Jia, Jiaya},
  journal={arXiv preprint arXiv:2505.12081},
  year={2025}
}

@article{liu2025seg,
  title={Seg-zero: Reasoning-chain guided segmentation via cognitive reinforcement},
  author={Liu, Yuqi and Peng, Bohao and Zhong, Zhisheng and Yue, Zihao and Lu, Fanbin and Yu, Bei and Jia, Jiaya},
  journal={arXiv preprint arXiv:2503.06520},
  year={2025}
}

@article{zhang2025chain,
  title={Chain-of-Focus: Adaptive Visual Search and Zooming for Multimodal Reasoning via RL},
  author={Zhang, Xintong and Gao, Zhi and Zhang, Bofei and Li, Pengxiang and Zhang, Xiaowen and Liu, Yang and Yuan, Tao and Wu, Yuwei and Jia, Yunde and Zhu, Song-Chun and others},
  journal={arXiv preprint arXiv:2505.15436},
  year={2025}
}

@article{jiang2025t2i,
  title={T2i-r1: Reinforcing image generation with collaborative semantic-level and token-level cot},
  author={Jiang, Dongzhi and Guo, Ziyu and Zhang, Renrui and Zong, Zhuofan and Li, Hao and Zhuo, Le and Yan, Shilin and Heng, Pheng-Ann and Li, Hongsheng},
  journal={arXiv preprint arXiv:2505.00703},
  year={2025}
}

@article{deng2025emerging,
  title={Emerging properties in unified multimodal pretraining},
  author={Deng, Chaorui and Zhu, Deyao and Li, Kunchang and Gou, Chenhui and Li, Feng and Wang, Zeyu and Zhong, Shu and Yu, Weihao and Nie, Xiaonan and Song, Ziang and others},
  journal={arXiv preprint arXiv:2505.14683},
  year={2025}
}

@article{duan2025got,
  title={Got-r1: Unleashing reasoning capability of mllm for visual generation with reinforcement learning},
  author={Duan, Chengqi and Fang, Rongyao and Wang, Yuqing and Wang, Kun and Huang, Linjiang and Zeng, Xingyu and Li, Hongsheng and Liu, Xihui},
  journal={arXiv preprint arXiv:2505.17022},
  year={2025}
}

@article{liu2023lost,
  title={Lost in the middle: How language models use long contexts},
  author={Liu, Nelson F and Lin, Kevin and Hewitt, John and Paranjape, Ashwin and Bevilacqua, Michele and Petroni, Fabio and Liang, Percy},
  journal={arXiv preprint arXiv:2307.03172},
  year={2023}
}

@article{xiao2023efficient,
  title={Efficient streaming language models with attention sinks},
  author={Xiao, Guangxuan and Tian, Yuandong and Chen, Beidi and Han, Song and Lewis, Mike},
  journal={arXiv preprint arXiv:2309.17453},
  year={2023}
}

@article{an2024make,
  title={Make your llm fully utilize the context},
  author={An, Shengnan and Ma, Zexiong and Lin, Zeqi and Zheng, Nanning and Lou, Jian-Guang and Chen, Weizhu},
  journal={Advances in Neural Information Processing Systems},
  volume={37},
  pages={62160--62188},
  year={2024}
}

@article{yu2024unveiling,
  title={Unveiling and harnessing hidden attention sinks: Enhancing large language models without training through attention calibration},
  author={Yu, Zhongzhi and Wang, Zheng and Fu, Yonggan and Shi, Huihong and Shaikh, Khalid and Lin, Yingyan Celine},
  journal={arXiv preprint arXiv:2406.15765},
  year={2024}
}

@article{yu2024mitigate,
  title={Mitigate position bias in large language models via scaling a single dimension},
  author={Yu, Yijiong and Jiang, Huiqiang and Luo, Xufang and Wu, Qianhui and Lin, Chin-Yew and Li, Dongsheng and Yang, Yuqing and Huang, Yongfeng and Qiu, Lili},
  journal={arXiv preprint arXiv:2406.02536},
  year={2024}
}

@inproceedings{
zhang2024tell,
title={Tell Your Model Where to Attend: Post-hoc Attention Steering for {LLM}s},
author={Qingru Zhang and Chandan Singh and Liyuan Liu and Xiaodong Liu and Bin Yu and Jianfeng Gao and Tuo Zhao},
booktitle={The Twelfth International Conference on Learning Representations},
year={2024},
url={https://openreview.net/forum?id=xZDWO0oejD}
}

@inproceedings{
chen2025why,
title={Why Is Spatial Reasoning Hard for {VLM}s? An Attention Mechanism Perspective on Focus Areas},
author={Shiqi Chen and Tongyao Zhu and Ruochen Zhou and Jinghan Zhang and Siyang Gao and Juan Carlos Niebles and Mor Geva and Junxian He and Jiajun Wu and Manling Li},
booktitle={Forty-second International Conference on Machine Learning},
year={2025},
url={https://openreview.net/forum?id=k7vcuqLK4X}
}

@misc{anthropic2024claude35,
  title = {Introducing Claude 3.5 Sonnet},
  author = {Anthropic},
  year = {2024},
  url = {https://www.anthropic.com/news/claude-3-5-sonnet},
  note = {Accessed: 2025-08-28}
}

@article{guardieiro2025instruction,
  title={Instruction Following by Boosting Attention of Large Language Models},
  author={Guardieiro, Vitoria and Stein, Adam and Khare, Avishree and Wong, Eric},
  journal={arXiv preprint arXiv:2506.13734},
  year={2025}
}

@article{venkateswaran2025spotlight,
  title={Spotlight Your Instructions: Instruction-following with Dynamic Attention Steering},
  author={Venkateswaran, Praveen and Contractor, Danish},
  journal={arXiv preprint arXiv:2505.12025},
  year={2025}
}

@article{enguehard2023sequential,
  title={Sequential integrated gradients: a simple but effective method for explaining language models},
  author={Enguehard, Joseph},
  journal={arXiv preprint arXiv:2305.15853},
  year={2023}
}

@inproceedings{
gong2025boosting,
title={Boosting the visual interpretability of {CLIP} via adversarial fine-tuning},
author={Shizhan Gong and Haoyu LEI and Qi Dou and Farzan Farnia},
booktitle={The Thirteenth International Conference on Learning Representations},
year={2025},
url={https://openreview.net/forum?id=khuIvzxPRp}
}

@article{zhang2024redundancy,
  title={From redundancy to relevance: Enhancing explainability in multimodal large language models},
  author={Zhang, Xiaofeng and Shen, Chen and Yuan, Xiaosong and Yan, Shaotian and Xie, Liang and Wang, Wenxiao and Gu, Chaochen and Tang, Hao and Ye, Jieping},
  journal={arXiv e-prints},
  pages={arXiv--2406},
  year={2024}
}

@inproceedings{ben2024lvlm,
  title={Lvlm-intrepret: An interpretability tool for large vision-language models},
  author={Ben Melech Stan, Gabriela and Aflalo, Estelle and Rohekar, Raanan Yehezkel and Bhiwandiwalla, Anahita and Tseng, Shao-Yen and Olson, Matthew Lyle and Gurwicz, Yaniv and Wu, Chenfei and Duan, Nan and Lal, Vasudev},
  booktitle={Proceedings of the IEEE/CVF Conference on Computer Vision and Pattern Recognition},
  pages={8182--8187},
  year={2024}
}

@article{zimmermann2025needle,
  title={Needle in a Patched Haystack: Evaluating Saliency Maps for Vision LLMs},
  author={Zimmermann, Bastien and Boussard, Matthieu and Craft, AI},
  journal={CNIA},
  pages={147},
  year={2025}
}

@inproceedings{zhou2016learning,
  title={Learning deep features for discriminative localization},
  author={Zhou, Bolei and Khosla, Aditya and Lapedriza, Agata and Oliva, Aude and Torralba, Antonio},
  booktitle={Proceedings of the IEEE conference on computer vision and pattern recognition},
  pages={2921--2929},
  year={2016}
}

@inproceedings{selvaraju2017grad,
  title={Grad-cam: Visual explanations from deep networks via gradient-based localization},
  author={Selvaraju, Ramprasaath R and Cogswell, Michael and Das, Abhishek and Vedantam, Ramakrishna and Parikh, Devi and Batra, Dhruv},
  booktitle={Proceedings of the IEEE international conference on computer vision},
  pages={618--626},
  year={2017}
}

@article{lapuschkin2019unmasking,
  title={Unmasking Clever Hans predictors and assessing what machines really learn},
  author={Lapuschkin, Sebastian and W{\"a}ldchen, Stephan and Binder, Alexander and Montavon, Gr{\'e}goire and Samek, Wojciech and M{\"u}ller, Klaus-Robert},
  journal={Nature communications},
  volume={10},
  number={1},
  pages={1096},
  year={2019},
  publisher={Nature Publishing Group UK London}
}

@inproceedings{abnar2020quantifying,
  title={Quantifying Attention Flow in Transformers},
  author={Abnar, Samira and Zuidema, Willem},
  booktitle={Proceedings of the 58th Annual Meeting of the Association for Computational Linguistics},
  year={2020},
  organization={Association for Computational Linguistics}
}

@inproceedings{sundararajan2017axiomatic,
  title={Axiomatic attribution for deep networks},
  author={Sundararajan, Mukund and Taly, Ankur and Yan, Qiqi},
  booktitle={International conference on machine learning},
  pages={3319--3328},
  year={2017},
  organization={PMLR}
}

@inproceedings{ribeiro2016should,
  title={" Why should i trust you?" Explaining the predictions of any classifier},
  author={Ribeiro, Marco Tulio and Singh, Sameer and Guestrin, Carlos},
  booktitle={Proceedings of the 22nd ACM SIGKDD international conference on knowledge discovery and data mining},
  pages={1135--1144},
  year={2016}
}

@incollection{NIPS2017_7062,
title = {A Unified Approach to Interpreting Model Predictions},
author = {Lundberg, Scott M and Lee, Su-In},
year = {2017},
publisher = {Curran Associates, Inc.},
url = {http://papers.nips.cc/paper/7062-a-unified-approach-to-interpreting-model-predictions.pdf}
}

@article{li2025token,
  title={Token Activation Map to Visually Explain Multimodal LLMs},
  author={Li, Yi and Wang, Hualiang and Ding, Xinpeng and Wang, Haonan and Li, Xiaomeng},
  journal={arXiv preprint arXiv:2506.23270},
  year={2025}
}

@article{shen2025glimpse,
  title={GLIMPSE: Gradient-Layer Importance Mapping for Prompted Visual Saliency Explanation for Generative LVLMs},
  author={Shen, Guanxi},
  journal={arXiv preprint arXiv:2506.18985},
  year={2025}
}

@article{vaswani2017attention,
  title={Attention is all you need},
  author={Vaswani, Ashish and Shazeer, Noam and Parmar, Niki and Uszkoreit, Jakob and Jones, Llion and Gomez, Aidan N and Kaiser, {\L}ukasz and Polosukhin, Illia},
  journal={Advances in neural information processing systems},
  volume={30},
  year={2017}
}

@article{dosovitskiy2020image,
  title={An image is worth 16x16 words: Transformers for image recognition at scale},
  author={Dosovitskiy, Alexey and Beyer, Lucas and Kolesnikov, Alexander and Weissenborn, Dirk and Zhai, Xiaohua and Unterthiner, Thomas and Dehghani, Mostafa and Minderer, Matthias and Heigold, Georg and Gelly, Sylvain and others},
  journal={arXiv preprint arXiv:2010.11929},
  year={2020}
}

@article{shi2024keep,
  title={Keep the cost down: A review on methods to optimize LLM's KV-cache consumption},
  author={Shi, Luohe and Zhang, Hongyi and Yao, Yao and Li, Zuchao and Zhao, Hai},
  journal={arXiv preprint arXiv:2407.18003},
  year={2024}
}

@inproceedings{
gandelsman2024interpreting,
title={Interpreting {CLIP}'s Image Representation via Text-Based Decomposition},
author={Yossi Gandelsman and Alexei A Efros and Jacob Steinhardt},
booktitle={The Twelfth International Conference on Learning Representations},
year={2024},
url={https://openreview.net/forum?id=5Ca9sSzuDp}
}

@article{yu2023neuron,
  title={Neuron-level knowledge attribution in large language models},
  author={Yu, Zeping and Ananiadou, Sophia},
  journal={arXiv preprint arXiv:2312.12141},
  year={2023}
}

@article{yang2025visionthink,
  title={VisionThink: Smart and Efficient Vision Language Model via Reinforcement Learning},
  author={Yang, Senqiao and Li, Junyi and Lai, Xin and Yu, Bei and Zhao, Hengshuang and Jia, Jiaya},
  journal={arXiv preprint arXiv:2507.13348},
  year={2025}
}

@article{bai2025qwen2,
  title={Qwen2. 5-vl technical report},
  author={Bai, Shuai and Chen, Keqin and Liu, Xuejing and Wang, Jialin and Ge, Wenbin and Song, Sibo and Dang, Kai and Wang, Peng and Wang, Shijie and Tang, Jun and others},
  journal={arXiv preprint arXiv:2502.13923},
  year={2025}
}

@article{zheng2024llamafactory,
  title={Llamafactory: Unified efficient fine-tuning of 100+ language models},
  author={Zheng, Yaowei and Zhang, Richong and Zhang, Junhao and Ye, Yanhan and Luo, Zheyan and Feng, Zhangchi and Ma, Yongqiang},
  journal={arXiv preprint arXiv:2403.13372},
  year={2024}
}

@misc{vonwerra2022trl,
  author = {Leandro von Werra and Younes Belkada and Lewis Tunstall and Edward Beeching and Tristan Thrush and Nathan Lambert and Shengyi Huang and Kashif Rasul and Quentin Gallouédec},
  title = {TRL: Transformer Reinforcement Learning},
  year = {2020},
  publisher = {GitHub},
  journal = {GitHub repository},
  howpublished = {\url{https://github.com/huggingface/trl}}
}

@article{hu2022lora,
  title={Lora: Low-rank adaptation of large language models.},
  author={Hu, Edward J and Shen, Yelong and Wallis, Phillip and Allen-Zhu, Zeyuan and Li, Yuanzhi and Wang, Shean and Wang, Lu and Chen, Weizhu and others},
  journal={ICLR},
  volume={1},
  number={2},
  pages={3},
  year={2022}
}

@article{achtibat2024attnlrp,
  title={Attnlrp: attention-aware layer-wise relevance propagation for transformers},
  author={Achtibat, Reduan and Hatefi, Sayed Mohammad Vakilzadeh and Dreyer, Maximilian and Jain, Aakriti and Wiegand, Thomas and Lapuschkin, Sebastian and Samek, Wojciech},
  journal={arXiv preprint arXiv:2402.05602},
  year={2024}
}

@article{chen2015microsoft,
  title={Microsoft coco captions: Data collection and evaluation server},
  author={Chen, Xinlei and Fang, Hao and Lin, Tsung-Yi and Vedantam, Ramakrishna and Gupta, Saurabh and Doll{\'a}r, Piotr and Zitnick, C Lawrence},
  journal={arXiv preprint arXiv:1504.00325},
  year={2015}
}

@inproceedings{rasheed2024glamm,
  title={Glamm: Pixel grounding large multimodal model},
  author={Rasheed, Hanoona and Maaz, Muhammad and Shaji, Sahal and Shaker, Abdelrahman and Khan, Salman and Cholakkal, Hisham and Anwer, Rao M and Xing, Eric and Yang, Ming-Hsuan and Khan, Fahad S},
  booktitle={Proceedings of the IEEE/CVF Conference on Computer Vision and Pattern Recognition},
  pages={13009--13018},
  year={2024}
}

@inproceedings{zhou2024openpsg,
  title={Openpsg: Open-set panoptic scene graph generation via large multimodal models},
  author={Zhou, Zijian and Zhu, Zheng and Caesar, Holger and Shi, Miaojing},
  booktitle={European Conference on Computer Vision},
  pages={199--215},
  year={2024},
  organization={Springer}
}

@article{chaoyou2023mme,
  title={Mme: A comprehensive evaluation benchmark for multimodal large language models},
  author={Chaoyou, Fu and Peixian, Chen and Yunhang, Shen and Yulei, Qin and Mengdan, Zhang and Xu, Lin and Jinrui, Yang and Xiawu, Zheng and Ke, Li and Xing, Sun and others},
  journal={arXiv preprint arXiv:2306.13394},
  volume={3},
  year={2023}
}

@article{li2023evaluating,
  title={Evaluating object hallucination in large vision-language models},
  author={Li, Yifan and Du, Yifan and Zhou, Kun and Wang, Jinpeng and Zhao, Wayne Xin and Wen, Ji-Rong},
  journal={arXiv preprint arXiv:2305.10355},
  year={2023}
}

@inproceedings{liu2024mmbench,
  title={Mmbench: Is your multi-modal model an all-around player?},
  author={Liu, Yuan and Duan, Haodong and Zhang, Yuanhan and Li, Bo and Zhang, Songyang and Zhao, Wangbo and Yuan, Yike and Wang, Jiaqi and He, Conghui and Liu, Ziwei and others},
  booktitle={European conference on computer vision},
  pages={216--233},
  year={2024},
  organization={Springer}
}

@article{yue2024mmmu-pro,
  title={Mmmu-pro: A more robust multi-discipline multimodal understanding benchmark},
  author={Yue, Xiang and Zheng, Tianyu and Ni, Yuansheng and Wang, Yubo and Zhang, Kai and Tong, Shengbang and Sun, Yuxuan and Yu, Botao and Zhang, Ge and Sun, Huan and others},
  journal={arXiv preprint arXiv:2409.02813},
  year={2024}
}

@article{zhang2024lmms,
  title={Lmms-eval: Reality check on the evaluation of large multimodal models},
  author={Zhang, Kaichen and Li, Bo and Zhang, Peiyuan and Pu, Fanyi and Cahyono, Joshua Adrian and Hu, Kairui and Liu, Shuai and Zhang, Yuanhan and Yang, Jingkang and Li, Chunyuan and others},
  journal={arXiv preprint arXiv:2407.12772},
  year={2024}
}

@article{hurst2024gpt,
  title={Gpt-4o system card},
  author={Hurst, Aaron and Lerer, Adam and Goucher, Adam P and Perelman, Adam and Ramesh, Aditya and Clark, Aidan and Ostrow, AJ and Welihinda, Akila and Hayes, Alan and Radford, Alec and others},
  journal={arXiv preprint arXiv:2410.21276},
  year={2024}
}

@article{team2024gemini,
  title={Gemini 1.5: Unlocking multimodal understanding across millions of tokens of context},
  author={Team, Gemini and Georgiev, Petko and Lei, Ving Ian and Burnell, Ryan and Bai, Libin and Gulati, Anmol and Tanzer, Garrett and Vincent, Damien and Pan, Zhufeng and Wang, Shibo and others},
  journal={arXiv preprint arXiv:2403.05530},
  year={2024}
}

@article{tong2024cambrian,
  title={Cambrian-1: A fully open, vision-centric exploration of multimodal llms},
  author={Tong, Peter and Brown, Ellis and Wu, Penghao and Woo, Sanghyun and IYER, Adithya Jairam Vedagiri and Akula, Sai Charitha and Yang, Shusheng and Yang, Jihan and Middepogu, Manoj and Wang, Ziteng and others},
  journal={Advances in Neural Information Processing Systems},
  volume={37},
  pages={87310--87356},
  year={2024}
}

@article{chen2024expanding,
  title={Expanding performance boundaries of open-source multimodal models with model, data, and test-time scaling},
  author={Chen, Zhe and Wang, Weiyun and Cao, Yue and Liu, Yangzhou and Gao, Zhangwei and Cui, Erfei and Zhu, Jinguo and Ye, Shenglong and Tian, Hao and Liu, Zhaoyang and others},
  journal={arXiv preprint arXiv:2412.05271},
  year={2024}
}

@article{li2024llava,
  title={Llava-onevision: Easy visual task transfer},
  author={Li, Bo and Zhang, Yuanhan and Guo, Dong and Zhang, Renrui and Li, Feng and Zhang, Hao and Zhang, Kaichen and Zhang, Peiyuan and Li, Yanwei and Liu, Ziwei and others},
  journal={arXiv preprint arXiv:2408.03326},
  year={2024}
}

@article{yao2024minicpm,
  title={Minicpm-v: A gpt-4v level mllm on your phone},
  author={Yao, Yuan and Yu, Tianyu and Zhang, Ao and Wang, Chongyi and Cui, Junbo and Zhu, Hongji and Cai, Tianchi and Li, Haoyu and Zhao, Weilin and He, Zhihui and others},
  journal={arXiv preprint arXiv:2408.01800},
  year={2024}
}

@article{zhang2024internlm,
  title={Internlm-xcomposer-2.5: A versatile large vision language model supporting long-contextual input and output},
  author={Zhang, Pan and Dong, Xiaoyi and Zang, Yuhang and Cao, Yuhang and Qian, Rui and Chen, Lin and Guo, Qipeng and Duan, Haodong and Wang, Bin and Ouyang, Linke and others},
  journal={arXiv preprint arXiv:2407.03320},
  year={2024}
}

@article{xu2024llava,
  title={Llava-cot: Let vision language models reason step-by-step},
  author={Xu, Guowei and Jin, Peng and Wu, Ziang and Li, Hao and Song, Yibing and Sun, Lichao and Yuan, Li},
  journal={arXiv preprint arXiv:2411.10440},
  year={2024}
}

@inproceedings{dong2025insight,
  title={Insight-v: Exploring long-chain visual reasoning with multimodal large language models},
  author={Dong, Yuhao and Liu, Zuyan and Sun, Hai-Long and Yang, Jingkang and Hu, Winston and Rao, Yongming and Liu, Ziwei},
  booktitle={Proceedings of the Computer Vision and Pattern Recognition Conference},
  pages={9062--9072},
  year={2025}
}

@article{barez2025chain,
  title={Chain-of-thought is not explainability},
  author={Barez, Fazl and Wu, Tung-Yu and Arcuschin, Iv{\'a}n and Lan, Michael and Wang, Vincent and Siegel, Noah and Collignon, Nicolas and Neo, Clement and Lee, Isabelle and Paren, Alasdair and others},
  journal={Preprint, alphaXiv},
  pages={v2},
  year={2025}
}

@misc{yu2025explainablechainofthoughtreasoningempirical,
      title={Explainable Chain-of-Thought Reasoning: An Empirical Analysis on State-Aware Reasoning Dynamics}, 
      author={Sheldon Yu and Yuxin Xiong and Junda Wu and Xintong Li and Tong Yu and Xiang Chen and Ritwik Sinha and Jingbo Shang and Julian McAuley},
      year={2025},
      eprint={2509.00190},
      archivePrefix={arXiv},
      primaryClass={cs.CL},
      url={https://arxiv.org/abs/2509.00190}, 
}

@article{zhang2024mme,
  title={Mme-realworld: Could your multimodal llm challenge high-resolution real-world scenarios that are difficult for humans?},
  author={Zhang, Yi-Fan and Zhang, Huanyu and Tian, Haochen and Fu, Chaoyou and Zhang, Shuangqing and Wu, Junfei and Li, Feng and Wang, Kun and Wen, Qingsong and Zhang, Zhang and others},
  journal={arXiv preprint arXiv:2408.13257},
  year={2024}
}

@article{chen2024we,
  title={Are we on the right way for evaluating large vision-language models?},
  author={Chen, Lin and Li, Jinsong and Dong, Xiaoyi and Zhang, Pan and Zang, Yuhang and Chen, Zehui and Duan, Haodong and Wang, Jiaqi and Qiao, Yu and Lin, Dahua and others},
  journal={Advances in Neural Information Processing Systems},
  volume={37},
  pages={27056--27087},
  year={2024}
}

@article{masry2022chartqa,
  title={Chartqa: A benchmark for question answering about charts with visual and logical reasoning},
  author={Masry, Ahmed and Long, Do Xuan and Tan, Jia Qing and Joty, Shafiq and Hoque, Enamul},
  journal={arXiv preprint arXiv:2203.10244},
  year={2022}
}

@article{saikh2022scienceqa,
  title={Scienceqa: A novel resource for question answering on scholarly articles},
  author={Saikh, Tanik and Ghosal, Tirthankar and Mittal, Amish and Ekbal, Asif and Bhattacharyya, Pushpak},
  journal={International Journal on Digital Libraries},
  volume={23},
  number={3},
  pages={289--301},
  year={2022},
  publisher={Springer}
}

@article{shahgir2024illusionvqa,
  title={Illusionvqa: A challenging optical illusion dataset for vision language models},
  author={Shahgir, Haz Sameen and Sayeed, Khondker Salman and Bhattacharjee, Abhik and Ahmad, Wasi Uddin and Dong, Yue and Shahriyar, Rifat},
  journal={arXiv preprint arXiv:2403.15952},
  year={2024}
}

@article{dahou2025vision,
  title={Vision-Language Models Can't See the Obvious},
  author={Dahou, Yasser and Huynh, Ngoc Dung and Le-Khac, Phuc H and Para, Wamiq Reyaz and Singh, Ankit and Narayan, Sanath},
  journal={arXiv preprint arXiv:2507.04741},
  year={2025}
}

@inproceedings{wang2020score,
  title={Score-CAM: Score-weighted visual explanations for convolutional neural networks},
  author={Wang, Haofan and Wang, Zifan and Du, Mengnan and Yang, Fan and Zhang, Zijian and Ding, Sirui and Mardziel, Piotr and Hu, Xia},
  booktitle={Proceedings of the IEEE/CVF conference on computer vision and pattern recognition workshops},
  pages={24--25},
  year={2020}
}

@article{carion2025sam,
  title={Sam 3: Segment anything with concepts},
  author={Carion, Nicolas and Gustafson, Laura and Hu, Yuan-Ting and Debnath, Shoubhik and Hu, Ronghang and Suris, Didac and Ryali, Chaitanya and Alwala, Kalyan Vasudev and Khedr, Haitham and Huang, Andrew and others},
  journal={arXiv preprint arXiv:2511.16719},
  year={2025}
}

@inproceedings{tran2025reasonvqa,
  title={Reasonvqa: A multi-hop reasoning benchmark with structural knowledge for visual question answering},
  author={Tran, Duong T and Tran, Trung-Kien and Hauswirth, Manfred and Le Phuoc, Danh},
  booktitle={Proceedings of the IEEE/CVF International Conference on Computer Vision},
  pages={18793--18803},
  year={2025}
}

@article{lau2018dataset,
  title={A dataset of clinically generated visual questions and answers about radiology images},
  author={Lau, Jason J and Gayen, Soumya and Ben Abacha, Asma and Demner-Fushman, Dina},
  journal={Scientific data},
  volume={5},
  number={1},
  pages={180251},
  year={2018},
  publisher={Nature Publishing Group}
}

@inproceedings{xing2025re,
  title={Re-Align: Aligning vision language models via retrieval-augmented direct preference optimization},
  author={Xing, Shuo and Li, Peiran and Wang, Yuping and Bai, Ruizheng and Wang, Yueqi and Hu, Chan-Wei and Qian, Chengxuan and Yao, Huaxiu and Tu, Zhengzhong},
  booktitle={Proceedings of the 2025 Conference on Empirical Methods in Natural Language Processing},
  pages={2379--2397},
  year={2025}
}

@inproceedings{liu2025multi,
  title={Multi-Frequency Contrastive Decoding: Alleviating Hallucinations for Large Vision-Language Models},
  author={Liu, Bingqian and Zhang, Fu and Chen, Guoqing and Cheng, Jingwei},
  booktitle={Proceedings of the 2025 Conference on Empirical Methods in Natural Language Processing},
  pages={28556--28572},
  year={2025}
}

@article{li2025mitigating,
  title={Mitigating object hallucinations in mllms via multi-frequency perturbations},
  author={Li, Shuo and Sun, Jiajun and Zheng, Guodong and Fan, Xiaoran and Shen, Yujiong and Lu, Yi and Xi, Zhiheng and Yang, Yuming and Tan, Wenming and Ji, Tao and others},
  journal={arXiv preprint arXiv: 2503.14895},
  year={2025}
}
}

\clearpage
\setcounter{page}{1}
\maketitlesupplementary
\appendix
In this supplementary material, we first give the derivation of the saliency map techniques when the normalization layer is involved (Sec.~\ref{app:norm}). Then we provide statistics and curation details of our training data (Sec.~\ref{app:dataset}). We also give detailed implementation details (Sec.~\ref{app:imp}) as well as brief introductions of the benchmarks utilized (Sec.~\ref{app:bench}). Finally, we show the complete numerical results (Sec.~\ref{app:exp}).
\section{Normalization Layers}
\label{app:norm}
In this section, we outline the modifications required to incorporate normalizations layer (usually RMSNorm) into our method. Normalizations are applied in two key locations: before the projection layer (on the output of the transformer) and before each layer in the transformer block (on the layer input). We detail how the attribution techniques should be adjusted accordingly.

\subsection{Pre-unembedding Normalization}
In typical language models, a normalization layer, usually RMSNorm, is applied to the transformer's output before the unembedding layers. Formally, the final logits is given by $\text{RMSNorm}(\textbf{h}_i^L)\textbf{E}_u$, and the normalization layer can be written as:
\begin{equation}
\text{RMSNorm}(x)=\gamma\times\frac{x}{\sigma_l^2+\epsilon},
\end{equation}
where $x \in \mathbb{R}^d$ is the input token, $\sigma_l\in\mathbb{R}$ is the standard deviation, and $\gamma \in \mathbb{R}^d$ is learned vectors. As the normalization if a linear operation, when calculating the contribution of each visual tokens, we can directly apply the same operation to each token, i.e.,:
\begin{equation}
c_p=\sum_{l=1}^{L}\sum_{j=1}^H\text{RMSNorm}(\alpha_{i,j,p}^l{\mathbf{W}^l_{o{\color{black},j}}}{{\mathbf{W}^l_{v{\color{black},j}}}}\textbf{h}_p^{l-1})\textbf{E}_u.
\end{equation}
One major difference is that in this case, the standard deviation $\hat{\sigma}$ should be calculated from the final representation $\mathbf{h}_p^L$, i.e., $\hat{\sigma}=\text{s.d.}(\mathbf{h}_p^L)$.
\subsection{MSAs Input Layer Normalizations}
In the main paper, we do not describe the normalization layers that are applied to each input of the MSA block in the model. More accurately, the complete representation of the MSA block are:
\begin{equation}
\begin{split}
\tilde{\textbf{h}}_p^{l-1} &= \text{RMSNorm}(\textbf{h}_p^{l-1}),\\
\textbf{A}_i^l&=\sum_{j=1}^H\sum_{p=1}^i\alpha_{i,j,p}^l{\mathbf{W}^l_{o{\color{black},j}}}{{\mathbf{W}^l_{v{\color{black},j}}}}\tilde{\textbf{h}}_p^{l-1}, \\
\alpha^l_{i,j,p}&=\text{softmax}(\mathbf{W}^l_{q,j}\tilde{\textbf{h}}_i^{l-1}\cdot \mathbf{W}^l_{k,j}\tilde{\textbf{h}}_p^{l-1}).
\end{split}
\end{equation}
Therefore, by the linearity of the normalization layer, we can rewrite the token-wise contribution to incorporate the normalization:
\begin{equation}
c_p=\sum_{l=1}^{L}\sum_{j=1}^H\text{RMSNorm}(\alpha_{i,j,p}^l{\mathbf{W}^l_{o{\color{black},j}}}{{\mathbf{W}^l_{v{\color{black},j}}}}\tilde{\textbf{h}}_p^{l-1})\textbf{E}_u.
\end{equation}

\section{Description of Dataset}
\label{app:dataset}
In this section, we introduce the details of the dataset we used. The training of Saliency-R1 involves the use of two dataset, i.e., Saliency-R1-cold and Saliency-r1-8k.  
\subsection{Saliency-R1-cold}
Saliency-R1-cold is a subset of the Vision-R1-cold~\cite{huang2025vision}, which is a composition of LLaVA-CoT dataset (100K)~\cite{xu2024llava} and Mulberry dataset (200K)~\cite{yao2024mulberry}.  For samples with multi-round conversation, we split it and construct multiple single-round conversation sample. This is due the computational constraints that we cannot process text with extremely long contexts. Additionally, we filter out samples whose number of tokens in the response is greater than 1024. In the end, we have 272,881 samples. 

\subsection{Saliency-r1-8k}
Saliency-r1-8k is a subset of the Visual-CoT dataset~\cite{shao2024visual}, which is a composition of TextVQA, TextCaps, DocVQA, DUDE, SROIE, CUB-200-2011, flickr30k, Visual7W, InfographicsVQA, VSR, GQA, and Open images. We randomly sample 8,080 images from the Visual-CoT dataset and construct our Saliency-r1-8k dataset. The proportion of each sub-dataset remains the same as the original Visual-CoT dataset.

\section{Implementation Details}
In this section, we provide more details of the implementation for bot cold start and the saliency-map alignment.
\label{app:imp}
\subsection{SFT}
We use llama-factory~\cite{zheng2024llamafactory} as the framework for conducting SFT. We conduct SFT on Saliency-R1-cold for 2 epochs. The maximum learning rate is $5\times10^{-5}$ and then it is adjusted according to cosine annealing scheduler with warm-up ratio of 0.05. The batch size is 64. Lora with LoRA with rank 128 is used for optimization.  Other settings are set to be the default values of llama-factory.
\subsection{GRPO}
GRPO is performed using TRL as the framework. We conduct GRPO on Saliency-r1-8k for 3 epochs. We utilize an original learning rate of $1\times 10^{-5}$ and batch size of 128. LoRA of rank 8 is performed on the query and value projection layers. The number of rollouts is set to be 8. Other settings are set to be the default values of TRL.
\section{Introduction of benchmark}
\label{app:bench}
In this section, we provide a brief overview of the benchmarks used in this paper. Our experiments are conducted on well-established or recently introduced visual understanding benchmarks. These benchmarks were selected based on two key criteria: (1) The tasks must be vision-centric, meaning that the questions and answers should be derived from the content of the image, requiring attention to relevant regions within the image. Benchmarks that primarily focus on other aspects, such as mathematical reasoning or code generation, often depend heavily on the intrinsic reasoning abilities of the language model, which falls outside the scope of this paper. (2) The questions must be in a multiple-choice or True/False format, allowing us to directly evaluate the model's performance without being affected by the variability of LLM evaluation.

\noindent \textbf{MMMU-Pro.} MMMU-Pro~\cite{yue2024mmmu-pro} is a strengthened version of the MMMU benchmark that more rigorously tests multimodal understanding and reasoning. It filters out questions solvable by text-only LLMs, expands multiple-choice options up to 10 to curb option-guessing, and introduces a vision-only setting where questions and options are embedded in photos/screenshots to require simultaneous “seeing” and “reading.” 

\noindent\textbf{MMBench.} MMBench~\cite{liu2024mmbench} is a comprehensive benchmark for evaluating multimodal models’ understanding and reasoning abilities across fine-grained tasks. It spans a wide range of skills: visual perception, commonsense and logical reasoning, detail recognition, spatial relations, text recognition, and cross-modal alignment, primarily through carefully curated multiple-choice questions. With rigorous human verification and de-duplication to mitigate training data contamination, MMBench aims to provide a fair and reproducible assessment. 

\noindent \textbf{POPE.} POPE~\cite{li2023evaluating} is designed to evaluate object hallucination in vision–language models using binary-choice questions that verify the presence of specific objects in images. It reports metrics such as Accuracy, Recall, Precision, and F1 across three distinct sampling strategies, providing a robust and fine-grained assessment of hallucination tendencies. In our paper, we report the F1 score for POPE.

\noindent \textbf{MME.} The MME benchmark~\cite{chaoyou2023mme} evaluates multimodal models across 14 subtasks that span both perceptual processing and cognitive reasoning. By using carefully designed instruction–response pairs, MME reduces the likelihood of training data leakage, supporting a fair and rigorous assessment.

\noindent \textbf{MME-RealWorld.} MME-RealWorld~\cite{zhang2024mme} is a large, fully human-annotated benchmark for evaluating multimodal LLMs on high-resolution, real-world tasks. It comprises 13,366 images (avg $\sim$2000×1500) and 29,429 multiple-choice QA pairs across 43 subtasks in five domains: OCR in the wild, remote sensing, diagram/table understanding, video monitoring, and autonomous driving. Designed to stress fine-grained perception and reasoning, it includes an “E” option for unanswerable cases and enforces strict quality control by 25 annotators and 7 experts.

\noindent \textbf{MMStar.} MMStar~\cite{chen2024we} is a vision-indispensable multimodal benchmark of 1,500 human-curated, multiple-choice samples designed to fairly evaluate Large Vision-Language Models. It targets six core capabilities (coarse/fine-grained perception, instance/logical reasoning, science \& technology, mathematics) across 18 axes, with strict visual dependency, minimized data leakage, and balanced difficulty.

\noindent \textbf{ChartQA.} ChartQA~\cite{masry2022chartqa} is a benchmark for testing how effectively multimodal models answer questions about charts, emphasizing both visual comprehension and logical reasoning. It covers a range of chart types—such as bar charts and line graphs—and combines human-authored and automatically generated questions to probe complex reasoning. Importantly, ChartQA is highly OCR-dependent and demands fine-grained visual understanding, as models must extract textual information from charts and reason over it.

\noindent \textbf{IllusionVQA.} IllusionVQA~\cite{shahgir2024illusionvqa} is a two-part visual question answering benchmark designed to test Vision-Language Models on challenging optical illusions. It comprises 435 multiple-choice comprehension instances across 12 illusion categories curated from diverse real images, and a 1,000-sample “soft localization” task where models identify which of two side-by-side objects is geometrically impossible. In this paper, we mainly focus on the soft localization task.

\noindent \textbf{ScienceQA.} ScienceQA~\cite{saikh2022scienceqa} is a large-scale multimodal multiple-choice benchmark of about 21k K–12 science questions spanning natural, social, and language sciences, many with text and/or image contexts. Uniquely, it annotates most questions with “lectures” (background knowledge) and “explanations” (reasoning), enabling evaluation and training of chain-of-thought reasoning. The dataset covers 26 topics, 127 categories, and 379 skills, includes both diagrams and natural images, and supports both few-shot prompting and fine-tuning studies. 

\noindent \textbf{SalBench.} SalBench~\cite{dahou2025vision} is a vision-centric benchmark for Large Vision-Language Models that tests human-obvious visual saliency—finding an “odd-one-out” among distractors based on low-level features. Built by augmenting the P3 (synthetic) and O3 (natural) datasets with language instructions, it evaluates three tasks: Odd-One-Out Detection, Referring Odd-One-Out (text box coordinates), and Visual Referring Odd-One-Out (highlighted box). Features span color, orientation, size (synthetic) and extend to focus, shape, location, pattern (natural). 

\section{Additional Experimental Results}
\label{app:exp}
\subsection{Faithfulness Experiments}
In the main paper, we evaluate the faithfulness of the proposed saliency-map techniques on COCO Caption and OpenPSG datasets. Here we give additional experimental results on GranDf dataset (see Table~\ref{faithfulness_a}). The results further support the claim that the proposed saliency-map generation techniques achieve state-of-the-art faithfulness compared with previous saliency-map generation techniques.

In addition to the reported deletion and insertion metric, we conduct counterfactual analysis to show that Saliency-R1 would focus more on the foreground pixels compared with baselines. To do so, we inject the Gaussian noise of varied $\sigma$ to the foreground and background images of test set of saliency-r1-8k, and report the final accuracy. We compare the performance drop of the base model and Saliency-R1 in Table~\ref{faithfulness_b}. The results demonstrate that compared with the base model, the Saliency-R1 show larger performance drop when Gaussian noise is injected to the foreground pixels. On the other hand, with noisy background, the performance drop of Saliency-R1 is smaller than the base model. This means Saliency-R1 relies more on the foreground pixels compared with the base model. 

\begin{table}[tp]
\centering
\renewcommand\arraystretch{1}
\captionof{table}{\textbf{Faithfulness experiment results.} We use Qwen2.5-VL-3B-Instruct for experiments. Our saliency map technique achieves comparable or better faithfulness to SOTA methods regarding the deletion and insertion metrics. The best metric is \textbf{bold} and the second best metric is \underline{underlined}.}
\label{faithfulness_a}
\resizebox{0.5\textwidth}{!}{
\begin{tabular}{lcccccc}
\toprule
& \multicolumn{6}{c}{\textbf{GranDf}}\\
\cmidrule(lr){2-7}
\textbf{Method}&\multicolumn{3}{c}{\textbf{Deletion} $\downarrow$}&\multicolumn{3}{c}{\textbf{Insertion}  $\uparrow$}\\
\cmidrule(lr){2-4} \cmidrule(lr){5-7}
&\textbf{5\%}&\textbf{15\%}&\textbf{30\%}&\textbf{5\%}&\textbf{15\%}&\textbf{30\%}\\
\midrule
CAM~\citep{zhou2016learning}&81.71&78.21&74.14&-4.80&3.86&22.96\\
Grad-CAM~\citep{selvaraju2017grad}&81.71&78.21&74.14&-4.80&3.86&22.96\\
Raw Attention&79.72&73.02&66.58&-5.58&-2.06&15.91\\
Attention Rollout~\citep{abnar2020quantifying}&83.56&70.17&58.88&-4.94&1.36&25.98\\
ATTN-LRP~\citep{achtibat2024attnlrp}&70.97&57.06&\textbf{44.57}&-13.54&-1.58&30.38\\
TAM~\citep{li2025token}&78.40&73.45&66.89&\textbf{4.01}&\textbf{14.92}&\textbf{36.91}\\
\rowcolor{backcolour}\textbf{Ours}&\textbf{68.91}&\textbf{56.95}&\underline{49.22}&\underline{-0.59}&\underline{11.80}&\underline{36.64}\\
\bottomrule
\end{tabular}}
\end{table}
\begin{table}[tp]
\centering
\renewcommand\arraystretch{1}
\captionof{table}{\textbf{Counterfactual test results.} We inject Gaussian noise with different $\sigma$ to the foreground and background of the images, and prompt the model to answer visual questions. }
\label{faithfulness_b}
\resizebox{0.5\textwidth}{!}{
\begin{tabular}{lccccc}
\toprule
& \multicolumn{5}{c}{\textbf{Accuracy}}\\
\cmidrule(lr){2-6}
\textbf{Method}&\textbf{Clean}&\multicolumn{2}{c}{\textbf{Foreground} $\downarrow$}&\multicolumn{2}{c}{\textbf{Background} $\uparrow$}\\
 \cmidrule(lr){2-2} \cmidrule(lr){3-4} \cmidrule(lr){5-6}
&-&$\sigma=25$&$\sigma=50$&$\sigma=25$&$\sigma=50$\\
\midrule
Qwen2.5-VL-7B&70.8&60.8&63.9&50.5&55.4\\
\rowcolor{backcolour}Saliency-R1&74.4&58.0&66.4&48.9&59.0\\
\bottomrule
\end{tabular}}
\end{table}

\subsection{Robustness of Saliency-R1} We examine the robustness of Saliency-R1 on several additional dataset. 1) Noisy data: we inject Gaussian noise with different intensity into POPE and MME images, and evaluate the accuracy of the models with the noisy images as inputs 2) OOD data: we use VQA-RAD~\cite{lau2018dataset}, a medical VQA dataset with significant domain gap with the saliency-r1-8k, to show that Saliency-R1 can generalize well to the medical domain. 3) ReasonVQA~\cite{tran2025reasonvqa}: a multi-hop benchmark that requires the model to attend to multiple distinct regions across multiple image areas to correctly answer the question. Due to the huge size of the original validation set, we only sample 500 questions from the test set. We compare the performance of Saliency-R1 with base model, the results are shown in Table~\ref{robustness}. Saliency-R1 performs better than base model on these datasets, showing the robustness of the proposed method. 
\begin{table}[tp]
\centering
\renewcommand\arraystretch{1}
\captionof{table}{\textbf{Robustness Benchmark.} We compare Saliency-R1 with base model on several benchmarks that show the robustness and generalizability of the method. We inject Gaussian noise with different $\sigma$ to the images of POPE and MME.}
\label{robustness}
\resizebox{0.5\textwidth}{!}{
\begin{tabular}{lcccccc}
\toprule
\textbf{Method}&\makecell{\textbf{POPE} \\ $\sigma=25$}&\makecell{\textbf{POPE} \\ $\sigma=50$}&\makecell{\textbf{MME} \\ $\sigma=25$}&\makecell{\textbf{MME} \\ $\sigma=50$}&\textbf{VQA-RAD}&\textbf{ReasonVQA}\\
\midrule
Qwen2.5-VL-7B&85.1&81.1&2317&2223&73.1&42.2\\
\rowcolor{backcolour}Saliency-R1&86.9&83.2&2360&2252&75.4&56.9\\
\bottomrule
\end{tabular}}
\end{table}
\subsection{Throughput Analysis}
In Table~\ref{throughput}, we report the salieny-map generation time (including the inference time). All evaluations are conducted on a single NVIDIA A6000 GPU. The results demonstrate that our method maintains high training efficiency.
\begin{table}[tp]
\centering
\renewcommand\arraystretch{1}
\captionof{table}{\textbf{Throughput analysis.} We report the average time to generate one saliency map using the saliency-r1-8k dataset.}
\label{throughput}
\resizebox{0.5\textwidth}{!}{
\begin{tabular}{lcccccc}
\toprule
\textbf{Method}&CAM&Attention Rollout&ATTN-LRP&TAM & Ours\\
\midrule
s/samples& 2.29&3.89&4.52&5.56&2.59\\
\bottomrule
\end{tabular}}
\end{table}
\subsection{Comparison with Additional Hallucination-reduction Baselines.} We compare Saliency-R1 with three additional hallucination-reduction baselines on the POPE benchmark, namely RE-ALIGN~\cite{xing2025re}, MFCD~\cite{liu2025multi}, and MFP~\cite{li2025mitigating}. The results are shown in Table~\ref{hallucination}. Saliency-R1 performs favorably, achieving the highest performance among the evaluated methods.
\begin{table}[tp]
\centering
\renewcommand\arraystretch{1}
\captionof{table}{\textbf{Comparison with hallucination-reduction baselines.}}
\label{hallucination}
\begin{tabular}{lc}
\toprule
\textbf{Method}&\textbf{POPE}\\
\midrule
MFP-3B & 86.5 \\
RE-ALIGN-7B & 87.1 \\
MFCD-7B & 87.4 \\
\rowcolor{backcolour}Saliency-R1-3B & 87.5 \\
\rowcolor{backcolour}Saliency-R1-7B & 88.1 \\
\bottomrule
\end{tabular}
\end{table}

\subsection{Human Evaluation} To further show the improved interpretability of Saliency-R1, we conduct a small-scale user study. Specifically, we collect 10 VQA examples with the corresponding saliency maps, and ask 12 voluntters to score them on a scale of 1 to 5 regarding the quality of the interpretation. A saliency map of high quality should be visually coherent, relevant to the asked questions, and related to the thinking process and final answers. The average score of the base model is 3.6 and that of Saliency-R1 is 4.5. Wilcoxon signed-rank test shows the $p<0.05$. This further validates that Saliency-R1 presents better interpretability.

\subsection{Qualitative Examples}
\begin{figure*}[t]
\centering
\includegraphics[width=1\textwidth]{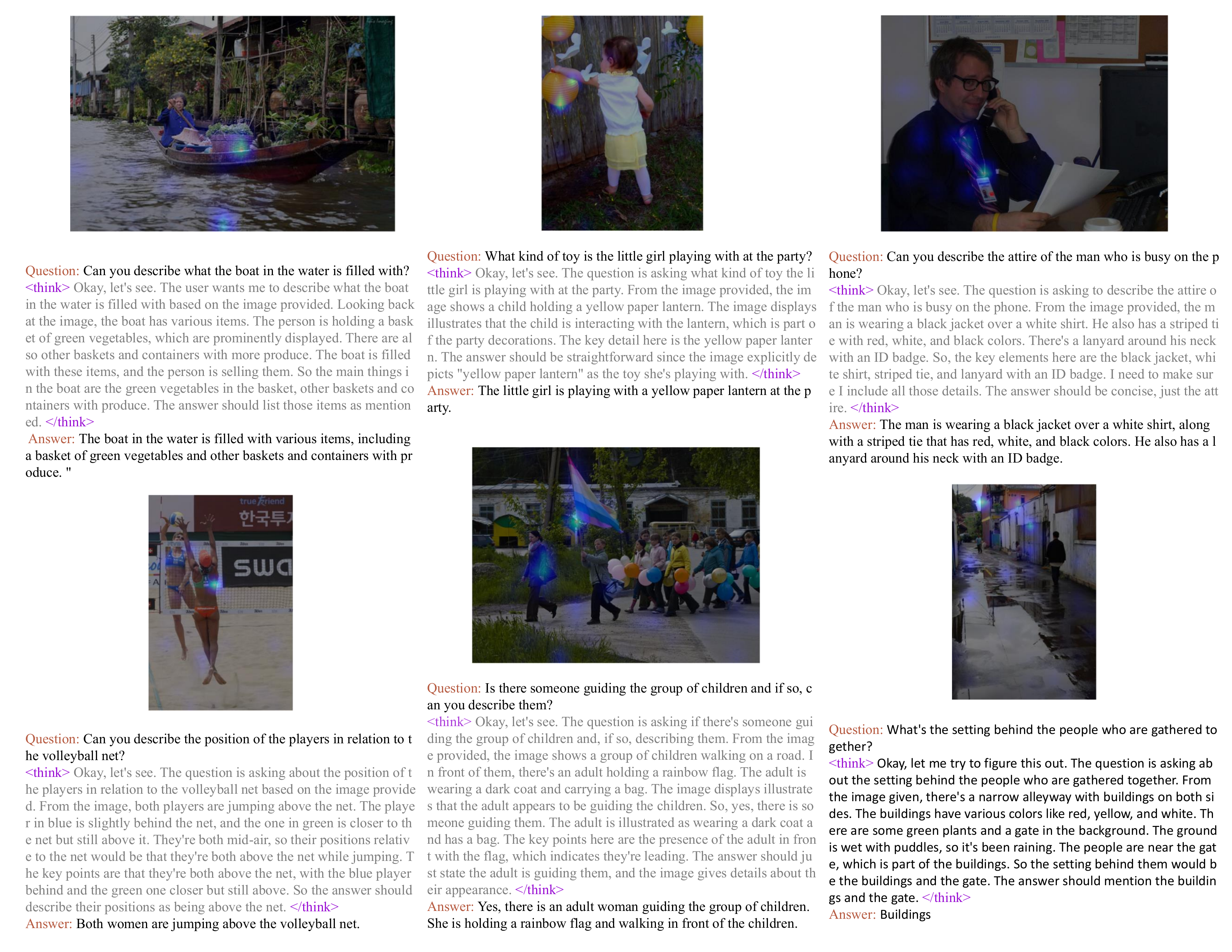}
\caption{Additional examples of the saliency maps generated by our proposed saliency map techniques, and the corresponding questions and responses. The examples are generated using Saliency-R1-7B.}
\vspace{-3mm}
\label{example_a}
\end{figure*}

In this section, we show more examples of the saliency maps generated to reflect the reasoning process of the model (see Fig.~\ref{example_a}). Our saliency map techniques can faithfully highlight the critical regions relevant to the questions.

\subsection{Complete Response in Fig.~\ref{example}}
In Fig.~\ref{example} of the main text, we present several examples of generated responses along with their corresponding saliency maps. Due to space limitations, we could not include the full thinking processes for some cases. Here, we provide the complete versions of these reasoning processes. From the examples, it is evident that the original Qwen2.5-VL model struggles with reasoning, often producing trivial content. After applying SFT, the model's reasoning capability improves significantly. However, we observe that the reasoning process often becomes excessively long, containing a substantial amount of irrelevant and unnecessary content. This irrelevant information dilutes the attention, causing the saliency map to highlight unrelated regions of the image. With saliency-map alignment, the reasoning becomes more concise and targeted. This not only improves the model's reasoning efficiency but also enhances its overall accuracy.

\noindent \textbf{Row 1.}
\begin{itemize}
\item \textbf{Qwen2.5-VL-7B.} The primary topic of the book is Cisco routers and switch management, as indicated by the title "Cisco Routers for the Desperate 2nd Edition." It suggests that the book is designed to help readers manage router and switch systems effectively, possibly in an easy or simplified manner.
\item \textbf{Saliency-R1-CI-7B.} Okay, let's see. The user provided a book cover titled "Cisco Routers for the Desperate: 2nd Edition." The subtitle displays router and switch management made easy. The author is Michael W. Lucas. The image has hands working on a Cisco router. The question is asking for the primary topic of the book based on the title. First, the main title is "Cisco Routers for the Desperate." The word "Cisco" here refers to the company known for networking equipment, specifically routers. The phrase "for the Desperate" suggests that the book is aimed at people who might be struggling or need help with their Cisco routers. The subtitle reinforces this by mentioning router and switch management. So putting it all together, the primary topic would be teaching readers how to manage Cisco routers and switches effectively. Even though there's an illustration of hands working on the router, the title itself points directly to Cisco routers. The subtitle adds more detail about the content being about management, but the main subject is still Cisco routers. Therefore, the primary topic is managing Cisco routers.
\item \textbf{Saliency-R1-7B.} Okay, let's see. The user provided a book cover titled "Cisco Routers for the Desperate: 2nd Edition." The question is asking for the primary topic of the book based on the title. Looking at the title, the main subject is clearly "Cisco Routers." The subtitle shows management, but the primary topic is the routers themselves. The image also displays the cover has an illustration of hands working on a router, reinforcing that the focus is on Cisco routers. So the primary topic is Cisco Routers.
\end{itemize}

\noindent \textbf{Row 2.}
\begin{itemize}
\item \textbf{Qwen2.5-VL-7B.} The chairs appear to be arranged around tables in the foreground of the image. They are positioned facing towards the front where the speaker is standing, suggesting that this is a setup for a presentation or lecture. The chairs are likely located on the left side of the image, as they are visible in the foreground and seem to be part of the audience seating area.
\item \textbf{Saliency-R1-CI-7B.} Okay, let's see. The question is asking where the chairs are located based on the image provided. First, I need to recall the details from the image. The image shows a conference room or meeting space. There's a stage with a podium where someone is speaking. In front of the stage, there are rows of tables arranged in an auditorium-style setup. Each table has chairs around them. The chairs are positioned directly in front of each table, facing the stage. So, the key points here are that the chairs are around the tables, which are set up in rows facing the stage. The image explicitly displays that the chairs are "positioned directly in front of each table, facing the stage." That means when people sit in the chairs, they're facing towards the stage where the speaker is standing. Therefore, the chairs are placed in front of the tables, which themselves are arranged in rows facing the stage. The answer should be straightforward since the image clearly illustrates their location relative to the tables and the stage.
\item \textbf{Saliency-R1-7B.} Okay, let's see. The question is asking where the chair is at based on the image provided. The image shows displays that the chairs are arranged around tables in rows, and they're positioned near the tables. So the chairs are at the tables. The answer should be straightforward from the given details.
\end{itemize}
\begin{table*}[t]
  \centering
  \caption{\textbf{Full Results of Ablation  Studies.} The best metric is \textbf{bold} and the second best is \underline{underlined}.}
  \begin{adjustbox}{width=\textwidth}
  \renewcommand{\arraystretch}{1.5}
  \setlength{\tabcolsep}{2pt}
  \begin{tabular}{@{}lcccccccccc@{}}
    \toprule
    \multirow{2}{*}{\textbf{Method}} & \textbf{MMMU-Pro} & \textbf{MMBench} & \textbf{POPE} & \textbf{MME} & \textbf{MME-RW
} & \textbf{MMStar}&\textbf{ChartQA}&\textbf{IllusionVQA}&\textbf{ScienceQA}&\textbf{SalBench}\\ 
    \cline{2-11}
     & test & en\_test & test & test & test & test & test & loc\_test & img\_test & p3\_test\\
    \midrule
    Qwen2.5-VL-7B~\citep{bai2025qwen2} & 36.2 & \underline{82.8} & 86.7 & 2302 & 58.7 & 62.4 & 84.0 & 37.5 & 88.2 & 49.1 \\
    Salieny-R1-zero&\textbf{37.8}&\textbf{82.9}&86.6&2279&60.6&60.0&75.2&35.7&86.6&53.6\\
Vision-R1&36.3&80.2&87.3&2371&62.3&62.0&87.9&37.8&94.0&56.5\\
Saliency-R1-attn&35.9&81.7&\textbf{88.3}&2377&62.6&62.1&87.6&\textbf{38.4}&93.7&60.3\\
Saliency-R1-pure&37.5&80.9&87.9&\textbf{2390}&\underline{62.7}&62.5&\textbf{88.2}&36.7&94.0&\underline{61.5}\\
Saliency-R1-think&36.1&82.3&82.4&2328&\underline{62.5}&61.1&86.6&35.3&\textbf{94.5}&58.5\\
Saliency-R1-answer&36.5&80.3&87.9&2378&62.0&60.6&87.8&35.7&93.1&55.2\\
     \rowcolor{backcolour} Saliency-R1 & \underline{37.6} & 81.8 & \underline{88.1} & \underline{2385} & \textbf{62.9} & \textbf{62.6} & \textbf{88.2} & \textbf{38.4} & \underline{94.3} & \textbf{63.7} \\
    \bottomrule
  \end{tabular}
  \end{adjustbox}
  \label{table:ablation}
\end{table*}
\noindent \textbf{Row 3.}
\begin{itemize}
\item \textbf{Qwen2.5-VL-7B.} The image shows a group of people gathered outdoors near a waterfront. There is no visible instrument or device to the left of any girl in the image. The focus seems to be on the people and their activities rather than any specific objects or devices.
\item \textbf{Saliency-R1-CI-7B.} Okay, let's break this down. The question is asking: "What kind of device is to the left of the instrument that the girl is to the right of?" First, I need to parse the image given. The image has a group near a waterfront with modern buildings. A girl is wearing a helmet and holding a guitar case. To her left, there's a device on the ground that looks like a camera or phone. The key here is understanding the spatial relationships. The girl is to the right of an instrument. The instrument mentioned is likely the guitar she's holding, since she's holding a guitar case. So the instrument in question is the guitar. Now, the device to the left of this guitar would be the one displayed as being on the ground to her left. According to the image, that device is a camera or phone. The question is asking for the type of device. Since both options are possible (camera or phone), but the image doesn't specify which, just depicts "camera or phone," the answer should reflect that ambiguity. However, the question might expect the more common association. But the original image shows displays "device resembling a camera or phone." So the correct answer would be either camera or phone. But since the question uses "device" without specifying, maybe the answer is "camera" or "phone." Wait, the problem illustrates "what kind of device," so the answer should be the specific type. But the image only specifies "camera or phone." So perhaps the answer is "camera" or "phone." But the question is phrased as "what kind of device," implying a singular answer. However, the image doesn't specify which one; it's either a camera or a phone. Therefore, the answer must be that it's a camera or phone. But the way the question is structured, maybe the answer is "camera" because that's more commonly associated with devices found on the ground at events. Alternatively, maybe the answer is "phone" if the context suggests a mobile device. But based strictly on the image provided, the answer is "camera or phone." However, the question might be expecting the most likely one. But since the instruction shows to provide the final answer without explanation, and the image shows displays "camera or phone," the answer should be those two possibilities. But the question is phrased as "what kind of device," so the answer is "camera or phone."
\item \textbf{Saliency-R1-7B.} Okay, let's break this down. The question is asking what kind of device is to the left of the instrument that the girl is to the right of. First, from the image, there's a girl with a guitar on her back. To the left of the guitar (which she's carrying), there's a device. The image shows illustrates a camera or similar device. So the device to the left of the guitar is a camera. The girl is to the right of the guitar, so the device is to the left of the guitar. The answer should be camera.
\end{itemize}

\subsection{Complete Experimental Results of Ablation Studies}
Here, we present the complete results of the ablation study. In addition to the experiments discussed in the main text, we conducted an additional experiment to evaluate the effectiveness of the proposed attention-rollout mechanism. Similar to the approach of Saliency-R1-think, we also tested a version of saliency maps that directly aggregates the answer tokens' saliency maps without applying attention rollout, referred to as Saliency-R1-answer. The results of this experiment are shown in Table~\ref{table:ablation}. Among all the ablation variants, the full version of Saliency-R1 demonstrates the best overall performance, highlighting the effectiveness of our design choices.

\end{document}